%% file: main_arxiv.tex
\documentclass[conference]{IEEEtran}
\IEEEoverridecommandlockouts

\usepackage[T1]{fontenc}
\usepackage[utf8]{inputenc}
\usepackage{amsmath,amssymb}
\usepackage{booktabs}
\usepackage{array}
\usepackage{multirow}
\usepackage{graphicx}
\usepackage{xcolor}
\usepackage{url}
\usepackage{cite}
\usepackage[hidelinks]{hyperref}

\newcommand{\nmult}{n_{\mathrm{mult}}}
\newcommand{\pjoint}{P_{\mathrm{joint}}}
\newcommand{\pindep}{P_{\mathrm{indep}}}
\newcommand{\code}[1]{\texttt{#1}}
\newcommand{\artifactkind}{a public}
\newcommand{\artifacturl}{\url{https://github.com/chenglin1112/evaluate-the-stack}}

\usepackage{orcidlink}
\hypersetup{pdftitle={Evaluate the Stack, Not the Layer: Do Deterministic and LLM Gates for Agent Actions Fail Independently?},pdfauthor={Chenglin Yang}}
\begin{document}

\title{Evaluate the Stack, Not the Layer:\\
       Do Deterministic and LLM Gates for Agent Actions Fail Independently?}

\author{\IEEEauthorblockN{Chenglin Yang\,\orcidlink{0009-0000-7241-1093}}
        \IEEEauthorblockA{University of Lancashire\\
        yangchenglin802@gmail.com}}

\maketitle

\input{sections/00_abstract}
\input{sections/01_introduction}

\input{sections/02_setup}
\input{sections/03_instrument}
\input{sections/04_instrument_fails}
\input{sections/05_readings}
\input{sections/06_replication}
\input{sections/07_related_work}
\input{sections/08_discussion}
\input{sections/09_limitations}
\input{sections/10_conclusion}

\input{sections/90_open_science_arxiv}
\input{sections/91_llm_usage}
\input{sections/92_ethics}

\bibliographystyle{IEEEtran}
\bibliography{references}

\appendices
\input{sections/99_appendix}

\end{document}

%% file: sections/00_abstract.tex
\begin{abstract}
Runtime gates for agent tool calls are stacked on the assumption that their errors
multiply. We test it on 1{,}119 labelled agent actions from three corpora, without an adaptive adversary. The stack has one deterministic rule layer and four LLM judges, three of them re-collected with the served model recorded on every call. We read each stack as a number of multiplication-equivalent layers, $\nmult$, with
its floor under perfect coupling. Under the STRICT miss definition (escalation to a human scored as
not stopped), any two judges compose to about 1.2 to 1.4 layers ($\varphi$ median $+0.430$, 6 of 6 pairs significant, floors 1.02 to
1.17).
The rule layer plus one judge composes to 1.86 to 2.09 layers ($\varphi$ median $+0.014$,
0 of 4 significant, floors 1.01 to 1.09). Under PRIMARY (escalation scored
as caught) the bands are 1.21 to 1.57 and 1.80 to
2.13. On the pooled data the point estimates separate, and so do the STRICT bootstrap
intervals; the exact intervals overlap in 2 of 24 rule-pair against judge-pair
combinations (10 of 24 under PRIMARY). On each corpus the rule pairs read above the judge pairs, on the internal
corpus only as censored lower bounds, and a version-pinned re-collection and a
third-vendor judge reproduce the split.
On a third-party corpus on which every layer misses most cases, the rule pairs sit
only midway between full coupling and independence before stratification.
Solo accuracy does not predict what a layer adds: a cloud rule pack lowers the rule
layer's solo miss rate by 20\% and adds no new joint coverage. The difficulty share of judge coupling is not identifiable: 31.8\% to 61.8\%
depending on the probe and the miss definition. One judge tier was served by an unrequested model version in 50 of 112 batches,
concentrated on the external corpus. That event overturned a pre-declared analysis rule, and the scoring of review verdicts reversed five conclusions. We report both.
\end{abstract}


%% file: sections/01_introduction.tex
\section{Introduction}\label{sec:intro}

An agent with tool access can delete a production database or push a credential to
a public repository in a single call \cite{register2025replit}. Deployments
therefore gate each proposed action before it executes. A common stack has a
deterministic rule layer that matches patterns in the action text, one or more LLM
judges that read the action and return a verdict, and sometimes a human reviewer
\cite{agenttrust2026a,fokou2026parallax}. The stack is justified by a multiplication
argument. If the rule layer misses 13.4\% of the actions labelled block in our
corpora and one judge misses 10.5\%, independent errors would give a joint miss rate
near 1.4\%. The safety
question this paper answers is whether stacked action gates have the independence
that argument assumes, and what each layer adds once the stack is measured as a
whole. The measurement is over labelled corpora of authored actions, without an
adaptive adversary; Section~\ref{sec:threat-model} states what that covers.

On our corpora the argument holds for one kind of pair and fails for another. The
rule layer and the judge above miss 1.3\% of block cases together (7 of 560), close
to the multiplication prediction. Two judges that each miss 10.5\% alone would be predicted to miss 1.1\%
together; they miss 5.2\% (29 of 560). These rates use the PRIMARY miss definition of
Section~\ref{sec:setup}. The difference between the two pairs is the subject of
this paper.

Failure correlation inside a defence stack has been measured before. The one prior study we found measured it for a seven-layer single-turn jailbreak-defence stack in which every member is a function of one target model \cite{alotaibi2026layered}. It describes
its scope as single-turn jailbreak only, excludes injection-targeted rows from its
measurement by design, lists correlation across agentic settings as unmeasured, and
collects its correlations under an adaptive adversary. Action gates differ in two ways that bear on correlation. A rule layer over action text consumes no model output, and the judges are black-box models from several
vendors that receive the actions directly, in the assume-compromise setting
Parallax names \cite{fokou2026parallax}. We measure the same statistic, with the same difficulty stratifier, in the setting that study excludes and without its adversary.

The paper makes a three-part argument. First, it defines an instrument. The number
of multiplication-equivalent layers, $\nmult$, reads an observed joint miss rate as the number of independent layers that would produce it, each failing at the stack's geometric-mean miss rate (Section~\ref{sec:instrument}). Second, it shows that the
instrument in current use fails (Section~\ref{sec:instrument-fails}). Solo accuracy,
the number reported for a gate measured alone, does not predict what the gate adds
to a stack. Adding a cloud rule pack cuts the rule layer's solo miss rate from 0.139 to 0.111. It adds zero new joint coverage on top of three judges under either definition and moves the rule layer toward the judge band on $\nmult$. A
50-pattern blocklist with a solo miss rate of 0.857 still adds nine cases under
STRICT. Third, it reports what
the new instrument reads (Section~\ref{sec:readings}). Under STRICT, any two judges compose to 1.22 to 1.44 layers above floors of 1.02 to 1.17. The rule layer plus one judge composes to 1.86 to 2.09 layers above floors of 1.01 to 1.09. On the pooled data the
two bands separate in their point estimates and, under STRICT, in their bootstrap
intervals; on each corpus the rule pairs read above the judge pairs, on the internal corpus only as censored lower bounds, and the intervals overlap. Three comparison collections give the same structure: the archived panel that preceded this one, a version-pinned re-collection of the same three tiers, and a fifth judge from a third vendor. Two others, a panel collected with a project-memory index in the judges' system context and a single-case check against batched judging, show that the verdicts are stable (Section~\ref{sec:replication}).

Our data do not separate judge coupling from shared difficulty. We stratify each
judge pair by a leave-both-out difficulty score, the construction the prior study
uses. The test for a shared surface mechanism runs in the direction opposite to that hypothesis. The share of coupling attributable to difficulty ranges from 31.8\% to 61.8\% with the choice of difficulty probe and miss definition
(Section~\ref{sec:where-coupling-comes-from}).
Difficulty alone can produce coincident failures between independently built
components \cite{eckhardt1985theoretical,knight1986experimental}, and the residual
cannot be read as evidence of a shared mechanism.

Two evaluation conventions changed conclusions during this study. One judge returns
the verdict \emph{review} on 47 of 560 block cases, the rule layer on 3, and the other
three judges on 2 to 17. Whether review counts as a
catch or as a miss reversed five conclusions, each time in the direction that
favoured the deferring judge (Section~\ref{sec:verdict-alphabet}). Every result is therefore
reported under both accounting rules, PRIMARY and STRICT (Section~\ref{sec:setup}).
Separately, one re-collected
judge tier was served by a version other than the one requested in 50 of its 112
batches, 36 of them with no trace except the per-call served-model field. The change fell unevenly across the corpora (Section~\ref{sec:version-changes}). The
archived verdicts have no such field.

This paper contributes:
\begin{itemize}
  \item a measurement of failure correlation among action gates on 1{,}119 labelled
    actions from three corpora, with one rule layer and four judges under one
    case-text condition, three judges re-collected with per-call provenance, and
    both miss definitions reported throughout
    (Sections~\ref{sec:setup} and~\ref{sec:readings});
  \item the instrument $\nmult$ with its floor and normalised reading, positioned
    against the beta factor and the Kish effective sample size, and a
    two-dimensional reading of stack value, mechanism class by in-domain
    competence, whose empty corner is stated as a falsifiable prediction
    (Sections~\ref{sec:instrument} and~\ref{sec:instrument-fails});
  \item a difficulty-versus-mechanism decomposition with a second implementation,
    replicating the prior stack measurement's non-identifiability finding in the
    setting that study excludes (Section~\ref{sec:where-coupling-comes-from});
  \item two evaluation conventions that reverse conclusions, a primary judge panel
    with per-call provenance and system-level blinding, five comparison
    collections including a version-pinned re-collection and a fifth judge from a
    third vendor, and a run on a third-party corpus
    (Sections~\ref{sec:verdict-alphabet} and~\ref{sec:replication});
  \item \artifactkind{} artifact with the per-layer verdicts, the per-call provenance
    records, the analysis scripts, the pre-registration and audit summaries, including
    the conclusions withdrawn during analysis.
\end{itemize}
Sections~\ref{sec:related}, \ref{sec:discussion} and~\ref{sec:limitations-m4} give related work, discussion and limitations.

%% file: sections/02_setup.tex
\section{Threat Model and Setup}\label{sec:setup}

\subsection{Threat model and scope}\label{sec:threat-model}
The adversary controls the text of the proposed action. This is the
assume-compromise setting that Parallax names \cite{fokou2026parallax}: the agent is
treated as already compromised, and the gates are evaluated on the actions it
proposes. Each gate sees the action text and nothing
else. The measurement is over labelled corpora of authored actions, and no adversary
adapts to the gates during it. The only adversarial proxy is a set of obfuscation
flags read off the case text (Section~\ref{sec:where-coupling-comes-from}), and
they were not optimised against any layer. Every $\varphi$ and $\nmult$ in this paper
is therefore a distributional statement about these corpora. It is not a worst-case
statement. An attacker who knows the rules can write an action that none of them matches. The external corpus shows that a distribution shift alone, with no attacker, takes the rule layer's miss rate to 52.9\%. An attacker who controls the action text
can also target the judges through it, and adaptive attacks of this kind defeat
published defences against jailbreaks and prompt injection
\cite{nasr2025attacker,zhan2025adaptive}. Adaptive evasion is out of scope. The
closest prior stack measurement collected its correlations under an adaptive
adversary \cite{alotaibi2026layered}. This paper does not. A deployer facing an
adaptive attacker cannot take an $\nmult$ near 2 as protection against an attack
aimed at the joint miss set.

\subsection{Layers}
The stack under test has one deterministic rule layer and four LLM judges. The rule
layer is the default policy engine of the AgentTrust runtime \cite{agenttrust2026a}:
174 rules, each a regular expression over the text of the proposed action, returning
allow, warn, block or review. It
reads no model output. The four judges are chat models that receive the action and return the same
four-way verdict. One, recorded by the operator as GPT-5.5 and called gpt55, is an
archived column with no version string, timestamp or temperature. The other three
were re-collected for this paper from one vendor through a coding-assistant harness,
in a session with no project memory, with the served model string recorded on every call. Haiku was requested as the dated \code{claude-haiku-4-5-20251001}, and sonnet and opus as the aliases \code{claude-sonnet-5} and \code{claude-opus-5}. The opus tier was served by \code{claude-opus-4-8} on 50 of its 112 batches (Section~\ref{sec:version-changes}).
Section~\ref{sec:replication} compares this panel with the archived columns it
replaces. A fifth judge
from a third vendor, re-collected with the served model recorded on every call, is
reported separately in Section~\ref{sec:cross-vendor-arm} because it covers two of
the three corpora.

\subsection{Elicitation}
Every judge column in the headline numbers was collected under one case-text
condition, which we call \emph{no-tell}. The judge is prompted in the judge role,
shown the action type, tool, content and parameters, asked to rate five risk
dimensions, and asked for a verdict. The human-written description field is blanked.
The archived gpt55 column was collected in batches of 90 cases in identifier
order with no record of wrapper prompt, temperature, model version or date. The
three re-collected columns were collected in batches of 10 cases in a seeded random
order over all 1{,}119 cases, one call per batch. Their case blocks are byte-identical to the archived no-tell
files; the instruction header is not. The archived collection used one header per corpus. The re-collections share a single header that drops the archived
header's list of example threat patterns and adds a no-tools clause. The header states the batch size, so the last external batch, which has 9 cases, has a header of its own. In addition,
100 cases per tier were judged again one case per call as a check on batching
(Section~\ref{sec:batch-vs-single}). The rule layer's archived
verdicts read the description. Recomputing them with it blanked changes 2 of 560
block verdicts and no headline conclusion (Section~\ref{sec:limitations-m4},
item 11). The description matters because on block cases it often names the threat:
126 of the 560 block-case descriptions contain a giveaway
word. Two archived columns were elicited with
the description shown, which we call \emph{faithful}, namely a DeepSeek column and a
second Opus column. Both are excluded from every headline number, since mixing
conditions would confound coupling with elicitation. The exclusion is conservative: including the DeepSeek column raises the
significant judge pairs from 6 of 6 to 9 of 10 under PRIMARY.

\subsection{Corpora}
The 1{,}119 cases with complete coverage on all five layers come from three corpora
with disjoint identifier spaces (Table~\ref{tab:corpora}). The
internal corpus (300 cases) is the runtime's own benchmark, six categories of fifty
cases each, and served as the rule layer's tuning set \cite{agenttrust2026a}. The
independent corpus (630 cases) was authored by the project team of
\cite{agenttrust2026a} in five batches after the rule set of a frozen release. When a case exposed a missing rule, the rule was added to the production set and the case kept \cite{agenttrust2026a}. The rule layer measured here is the current default,
after those patches, so that corpus is independently constructed but not zero-shot for the rule layer. Its 6.1\% rule miss rate is a post-patch residual. Its categories are verdict tiers
such as safe, moderate and dangerous. The external corpus (189 cases) covers cloud
identity and secrets, CI and infrastructure-as-code, databases, network egress,
package supply chains, shell persistence, file configuration and a deliberately hard
benign observability family \cite{agenttrust2026b}. Its cases were constructed with an LLM and labelled against a rubric. Every retained case carries a label that an independent human annotator who is not an author of the runtime's reports confirmed by blind relabelling. Of the 224 source cases, all relabelled, 193 agreed and 31 disagreed ($\kappa = 0.817$ before filtering). The disagreements were discarded, and the frozen 189 are agreed cases after tell and leakage filtering \cite{agenttrust2026b}.
The default rule set does not target that domain, and
Section~\ref{sec:decoupling-not-usefulness} shows what that costs. Every case
carries a truth label of allow, warn or block.

\begin{table}[t]
\caption{Corpora and truth labels. Misses are measured on block cases and false
alarms on allow cases; warn cases belong to neither domain.}
\label{tab:corpora}
\centering
\begin{tabular}{lrrrr}
\toprule
corpus & cases & block & allow & warn \\
\midrule
internal & 300 & 129 & 133 & 38 \\
independent & 630 & 344 & 188 & 98 \\
external (held out) & 189 & 87 & 58 & 44 \\
\midrule
all & 1{,}119 & 560 & 379 & 180 \\
\bottomrule
\end{tabular}
\end{table}

\subsection{Error domains}
Misses compose conjunctively, since the stack fails only if every layer misses, so
the miss domain is the 560 block cases. False alarms compose disjunctively, since
any layer that fires raises the alarm, and are reported separately on the 379 allow
cases; the 180 warn cases belong to neither domain.

\subsection{Two miss definitions}
The verdict alphabet contains \emph{review}, an escalation to a human that is neither
an execution nor a stop. On the 560 block cases the layers return it 3 (rules), 47 (gpt55), 17 (haiku), 2
(opus) and 12 (sonnet) times. Two readings
of a miss differ only on it. PRIMARY reads a miss as ``the action executed'' and
counts allow and warn as misses. STRICT reads a miss as ``the action was not
stopped'' and counts allow, warn and review. Which reading is correct depends on
whether a human review layer exists downstream to absorb the escalation, and this
study has no data on such a layer. Both definitions are reported everywhere, and
where they diverge we say so. The divergence is a finding of its own
(Section~\ref{sec:verdict-alphabet}).

\subsection{Verification and withdrawn conclusions}
Six headline statistics were recomputed against an independent implementation
(Appendix~\ref{app:tables}); one of them, the difficulty share, gained its second
implementation after review. Seven candidate conclusions were withdrawn during analysis. The pre-registration and a summary of the audits are in the artifact.

%% file: sections/03_instrument.tex
\section{The Instrument: Multiplication-Equivalent Layers}\label{sec:instrument}

For a stack $S$ with per-layer miss rates $p_L$ and observed joint miss rate
$\pjoint$, we define
\begin{equation}
\nmult = \frac{\ln \pjoint}{\ln g}, \qquad
g = \Big(\prod_{L \in S} p_L\Big)^{1/|S|}.
\label{eq:nmult}
\end{equation}
Layers whose errors multiply give $\nmult = |S|$. The value under perfect coupling
is not 1 in general. Since $\pjoint \le \min_L p_L$,
\begin{equation}
\nmult \;\ge\; \frac{\ln \min_L p_L}{\ln g},
\label{eq:floor}
\end{equation}
with equality when the misses of the most accurate layer are a subset of every
other layer's. This floor equals 1 only when the marginals are equal, and it rises
with their spread. For the pooled rule pairs under STRICT it is 1.01 to 1.09. On the
external corpus, where the rule layer misses half the block cases, it reaches 1.54
to 1.73. We report the floor
beside every $\nmult$ and, where the comparison matters, the normalised reading
$(\nmult - \mathrm{floor})/(|S| - \mathrm{floor})$, which is 0 at perfect coupling
and 1 under independence. Values above $|S|$ indicate negative association and
occur for the rule pairs whose $\varphi$ is below zero. Confidence intervals
propagate the exact Clopper--Pearson interval on $\pjoint$
\cite{clopper1934confidence}. A case-level bootstrap gives narrower intervals for
the judge pairs and similar ones for the rule pairs, and Table~\ref{tab:nmult}
reports both. When a stack has no joint
miss at the sample size, $\nmult$ has no point estimate, only a lower bound from the upper limit on $\pjoint$. We mark such a stack as censored.

The construction is an effective count: given an observed aggregate, report the
number of idealised identical units that would produce it. Hill numbers in ecology
use it, with a community's diversity read as the size of a reference community of
equally abundant species
\cite{macarthur1965patterns,macarthur1972geographical,hill1973diversity,jost2006entropy,jost2007partitioning}.
We apply it to a joint failure probability: given $\pjoint$, how many independent
layers, each failing at the geometric-mean rate $g$, would produce it. The construction is established. We did not find this application in the common-cause-failure family of reliability engineering, which uses ratio parameters, or in the effective-sample-size family of statistics. We claim no new
mathematics. If a standard definition of this application exists, we would align
with it.

Two quantities share the vocabulary. The beta factor \cite{iec61508}, the standard
treatment of common-cause failure, is the conditional probability that a common
cause takes the other channels too. It is a ratio, it moves inversely to $\nmult$, and it does not say how many layers a stack is worth. The symbol $\beta$ also denotes judge miss rates in work we cite \cite{ray2026enforced}. Kish's effective
sample size \cite{kish1965survey}, $(\sum w_i)^2 / \sum w_i^2$ over a weighted
sample's weights, has been applied to LLM judge panels \cite{kohli2026ninejudges}. It is a different quantity about a different object, so we do not reuse that name.

The normalisation by $g$ is a deliberate choice. $g$ is the geometric mean of the
layers' own miss rates, so $\nmult$ does not reward a layer for being accurate: a
strong layer whose failures coincide with its partner's still scores low.
Section~\ref{sec:decoupling-not-usefulness} shows that a stack's leak rate is the
product of independence and in-domain competence. $\nmult$ reads the first factor,
though competence still enters it through the marginals and their floor, so the floor is reported with it.

Companion statistics are reported per pair. The correlation $\varphi$ is the
mean-square contingency coefficient on the two binary error indicators, identical to
Pearson's $r$ on 0/1 data. It is the headline statistic because the closest prior measurement reports it
\cite{alotaibi2026layered}. $\varphi$ is bounded by the marginals. For $p_1 \le p_2$
the attainable maximum is
\begin{equation}
\varphi_{\max} = \sqrt{\frac{p_1 (1 - p_2)}{p_2 (1 - p_1)}},
\label{eq:phimax}
\end{equation}
their Eq.~8. Our rule layer and judges have unequal miss rates, so
Table~\ref{tab:pairs} reports $\varphi/\varphi_{\max}$ beside raw
$\varphi$. The coupling ratio $c = P_{\mathrm{obs}} / \pindep$ and Cohen's $\kappa$
\cite{cohen1960kappa} are computed per pair by the scripts in the artifact. Pairwise significance uses Fisher's exact test with
Holm correction \cite{holm1979simple}; each table states the family over which the
correction runs.

%% file: sections/04_instrument_fails.tex
\section{The Instrument in Use Fails: Solo Accuracy Does Not Predict Stack Value}\label{sec:instrument-fails}

We test whether a gate's solo accuracy, the number usually reported for it,
predicts what the gate adds to a stack. Every increment below is measured the same way. Its new coverage is the number of block cases that a baseline stack of three judges leaks and the increment catches, summed over the four such baselines, under both miss definitions (560 block cases). Table~\ref{tab:increments} lists four increments.

\begin{table}[t]
\caption{What an increment adds to a three-judge stack (560 block cases). New
coverage is the number of baseline leaks the increment catches, summed over the four
three-judge baselines, with the per-baseline breakdown; b1 is a second, weaker rule
engine.}
\label{tab:increments}
\centering
\scriptsize
\setlength{\tabcolsep}{2pt}
\begin{tabular}{@{}llrlrl@{}}
\toprule
 & & \multicolumn{2}{c}{PRIMARY} & \multicolumn{2}{c}{STRICT} \\
increment & class & solo & new (per baseline) & solo & new (per baseline) \\
\midrule
rule layer & det.\ lexical & 0.134 & 31 (3/3/3/22) & 0.139 & 86 (17/19/22/28) \\
rule layer + pack & det.\ lexical & 0.100 & 31 (3/3/3/22) & 0.111 & 86 (17/19/22/28) \\
fourth judge & LLM semantic & 0.099 & 21 (0/0/0/21) & 0.133 & 22 (0/3/6/13) \\
b1 (50 patterns) & det.\ lexical & 0.857 & 2 (0/0/0/2) & 0.857 & 9 (2/3/2/2) \\
\bottomrule
\end{tabular}
\end{table}

Under STRICT the rule layer adds 86 cases against 22 for a fourth judge. Under
PRIMARY the two add 31 and 21, and 22 of the rule layer's 31 come from the one
baseline that lacks gpt55. The reason is the accounting rule for review. Under
PRIMARY gpt55's 47 review verdicts count as caught, so the three baselines that contain it leak only 5 cases each. The fourth judge added to the remaining baseline is gpt55 itself, credited with 8 deferrals among its 21 additions. Under
STRICT those baselines leak 19 to 25 cases and the rule layer catches 17 to 22 of
them, while a fourth judge catches 0 to 6. The rule layer's advantage is largest under STRICT, and the fourth
judge's value is one of the conclusions that the accounting rule moves
(Section~\ref{sec:verdict-alphabet}).

Two observations point in opposite directions, and both hold under either
definition. First, the cloud pack improves the rule layer's solo miss rate (0.139 to
0.111 under STRICT, a 20\% relative improvement) and leaves the stack's leak set unchanged: 86 to 86 and 31 to 31, identical case for case. Inspected
directly, every block case the pack newly catches on the external corpus (17 under
PRIMARY, 15 under STRICT) was already caught by all four judges. On the independent
corpus the pack changes three block verdicts: it newly catches two cases, one of
them caught by three of the four judges, and it moves one case from block to
review. None of the three is leaked by any three-judge baseline, so new joint
coverage is exactly zero. The pack was written against
separate held-out corpora and did not consult the external corpus scored here, so
the solo improvement is itself a held-out result.
Second, b1, the layer with the worst solo accuracy, is not the best contributor. It
adds 9 cases under STRICT and 2 under PRIMARY, against 86 and 31 for the stronger
deterministic layer.

The pack also moves the rule layer on the independence axis, and that move should
be read together with the zero. On the external corpus, where almost all of the pack's effect falls, the rule layer's $\varphi$
with the four judges rises from $+0.10$, $-0.01$, $+0.10$ and $+0.18$ to $+0.15$,
$+0.08$, $+0.15$ and $+0.27$ (PRIMARY), and its $\nmult$ with each judge falls from
1.75, 2.03, 1.75 and 1.68 to 1.61, 1.81, 1.61 and 1.51; under STRICT it falls from
1.96, 1.76, 2.05 and 1.75 to 1.71, 1.56, 1.88 and
1.57. The joint miss
counts do not change (1, 2, 1 and 3 under PRIMARY; 7, 7, 1 and 5 under STRICT). The
denominator moves and the numerator does not. The pack removes only rule misses that
the judges already caught, which is the difficulty mechanism of Section~\ref{sec:where-coupling-comes-from} acting on the rule layer. The rule layer's independence reading moves toward the judge band as its competence rises on
cases the judges find easy.

b1 cannot be used to attribute decoupling to the mechanism class. Its solo miss rate of 0.857 gives its pairs with the judges floors of 1.83 to 1.88
(Section~\ref{sec:instrument}), which leaves only 0.12 to 0.17 up to the value of
2 that independence would give. The observed $\nmult$ of 1.91 to 1.97 cannot be
read on that scale, and its $\varphi/\varphi_{\max}$ with the judges is 0.20 to
0.63, inside the judge band. b1 therefore appears in this paper only as a point about
increments. The attribution of decoupling to the mechanism class rests on the
default rule layer across three corpora, the comparison panels and the cross-vendor
column (Sections~\ref{sec:readings} and~\ref{sec:replication}).

\begin{figure*}[t]
  \centering
  \includegraphics[width=\textwidth]{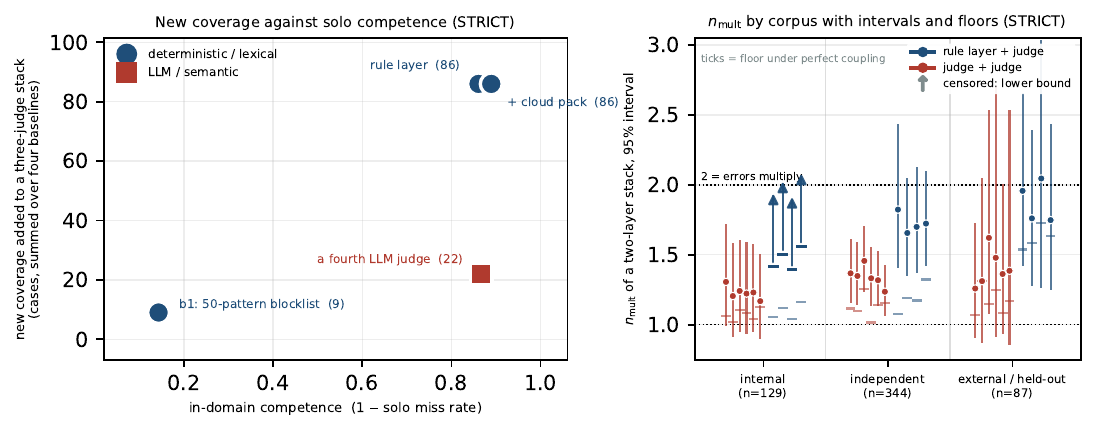}
  \caption{Left: new coverage added to a three-judge stack, summed over the four
  baselines (STRICT), against the increment's in-domain competence (one minus solo
  miss rate); marker shape gives the mechanism class. The empty corner of the
  $2 \times 2$, same mechanism with low competence, is a prediction of about zero
  added coverage, and no layer of that kind was available to test. Four points on
  three corners form a taxonomy consistent with the observations; they do not fit a
  two-factor model. Right: $\nmult$ of every two-layer stack by corpus (STRICT) with
  exact 95\% intervals. Short ticks mark each stack's floor under perfect coupling,
  which depends on the marginals. Censored stacks, with no joint miss at the sample
  size, are drawn as arrows from their lower bound; on the internal corpus every
  rule-plus-judge stack is censored. 560 block cases. Every judge column is no-tell;
  the rule layer's verdicts read the description field (Section~\ref{sec:setup}).}
  \label{fig:main}
\end{figure*}

Solo accuracy therefore over-predicts in one direction and under-predicts in the
other. The pattern consistent with all four points is two-dimensional: mechanism
class and in-domain competence both matter, and neither alone predicts stack value.
Four points on three corners of a $2 \times 2$ form a taxonomy consistent with the
observations. They do not fit a two-factor model, since two of the four points carry
the caveats above and there are too few points to separate the factors. The reading is an existence statement. The empty corner, same mechanism with low competence, is
a falsifiable prediction of about zero added coverage, and we had no such layer to
test. Figure~\ref{fig:main} (left) plots the four points in the units of Table~\ref{tab:increments}. The cloud-pack increment is horizontal on the coverage
axis and, on the independence axis of the right panel, a move toward the judge
band.

%% file: sections/05_readings.tex
\section{Read with the New Instrument: Coupling Splits Along Mechanism, Not Count}\label{sec:readings}

\subsection{Pairwise correlation}
Table~\ref{tab:pairs} gives the pairwise miss correlation on the pooled 560 block
cases. The rule layer's four pairs with judges sit near zero on raw $\varphi$ and stay
there after marginal normalisation: $\varphi/\varphi_{\max}$ median $+0.002$ under
PRIMARY and $+0.018$ under STRICT, against $+0.697$ and $+0.501$ for the six judge
pairs. The largest normalised rule pair
($+0.134$, rules with gpt55, PRIMARY) is below the smallest normalised judge pair
($+0.432$), so the split of point estimates does not depend on unequal marginals.
Significance is Fisher's exact test with Holm correction over the ten pairs of each
definition; normalisation rescales $\varphi$ and changes no $p$-value. Intervals add
one qualification. Under PRIMARY the 90\% intervals of the rule pairs are disjoint
from those of the three same-vendor judge pairs, whose lower limits are $+0.32$ or above. They overlap those of the pairs that involve gpt55, whose lower limits are
$+0.04$ to $+0.12$ because its review habit holds its $\varphi$ down
(Section~\ref{sec:verdict-alphabet}). Under
STRICT every judge pair has $\varphi \ge 0.328$ and the separation is complete,
with 6 of 6 judge pairs significant against 0 of 4 rule
pairs.

\begin{table*}[t]
\caption{Pairwise miss correlation, pooled ($n = 560$ block cases). Median and range
over the pairs in each group. Significance is Fisher's exact test with Holm
correction over the ten pairs of the same definition.}
\label{tab:pairs}
\centering
\small
\begin{tabular}{@{}lrlrlc@{}}
\toprule
group & \multicolumn{2}{c}{$\varphi$} & \multicolumn{2}{c}{$\varphi/\varphi_{\max}$} & sig. \\
 & med. & range & med. & range & \\
\midrule
\multicolumn{6}{@{}l}{PRIMARY} \\
rule $\times$ judge (4) & $+0.002$ & $[-0.038, +0.041]$ & $+0.002$ & $[-0.044, +0.134]$ & 0/4 \\
judge $\times$ judge (6) & $+0.342$ & $[+0.146, +0.604]$ & $+0.697$ & $[+0.432, +0.796]$ & 6/6 \\
\multicolumn{6}{@{}l}{STRICT} \\
rule $\times$ judge (4) & $+0.014$ & $[-0.025, +0.051]$ & $+0.018$ & $[-0.029, +0.052]$ & 0/4 \\
judge $\times$ judge (6) & $+0.430$ & $[+0.328, +0.603]$ & $+0.501$ & $[+0.429, +0.838]$ & 6/6 \\
\bottomrule
\end{tabular}
\end{table*}

The judge side reproduces the magnitude reported by the closest prior measurement,
$\varphi$ of 0.30 to 0.75 across the fifteen pairs of a seven-layer single-turn
jailbreak-defence stack whose every member is a function of one target model
\cite{alotaibi2026layered}. It is also consistent with the correlated errors reported across LLMs from different vendors \cite{goel2025alike,kim2025correlated}.
Stratifying by corpus with the Cochran--Mantel--Haenszel common odds ratio
\cite{mantel1959statistical} does not collapse the judge associations toward 1
(Appendix~\ref{app:tables}), so the pooled association is present within corpora. Pooling corpora of unequal difficulty does not create it. A Breslow--Day test finds
no heterogeneity of the per-corpus odds ratios for nine of the ten pairs
($p \ge 0.064$). The exception is gpt55 with opus ($p = 0.026$), whose opus column
was answered by different served versions in different corpora
(Section~\ref{sec:version-changes}), so a common odds ratio is not an admissible
summary for that pair. The rule pairs
have common odds ratios of 1.74 to 3.96, with $p$ from 0.12 to
0.50.

\subsection{Multiplication-equivalent layers}
Table~\ref{tab:nmult} gives $\nmult$ for every two-layer stack and for three larger
ones, with the floor of each stack and two intervals. Under STRICT, any two judges
compose to 1.22 to 1.44 layers above floors of 1.02 to 1.17 (same-vendor pairs 1.22
to 1.32, pairs with gpt55 1.34 to 1.44), with case-level bootstrap intervals spanning 1.16 to 1.55. The three same-vendor judges compose to 1.46 and all four judges to 1.65. The rule layer plus one judge composes to 1.86 to 2.09 layers above
floors of 1.01 to 1.09, with bootstrap intervals spanning 1.67 to 2.53. On the
normalised scale the rule pairs read 0.86 to 1.10 and the judge pairs 0.08 to
0.33. Rows with
five or fewer joint misses support bounds only; the two stacks with two joint
misses are censored in 13.7\% of bootstrap resamples
(Section~\ref{sec:limitations-m4}, item 2).

\begin{table*}[t]
\caption{$\nmult$ by stack ($n = 560$ block cases). $k$ is the number of joint
misses; rows with $k \le 5$ (marked $\dagger$) support bounds only. Floor is the
value under perfect coupling, Eq.~\eqref{eq:floor}. CP is the interval propagated
from the exact interval on $\pjoint$; boot is a case-level bootstrap interval
(4{,}000 resamples), reported for STRICT.}
\label{tab:nmult}
\centering
\footnotesize
\begin{tabular}{@{}lrrlrrll@{}}
\toprule
 & \multicolumn{3}{c}{PRIMARY} & \multicolumn{4}{c}{STRICT} \\
stack & floor & $\nmult$ ($k$) & CP 95\% & floor & $\nmult$ ($k$) & CP 95\% & boot 95\% \\
\midrule
haiku + opus & 1.00 & 1.32 (29) & [1.16, 1.49] & 1.05 & 1.32 (35) & [1.17, 1.48] & [1.24, 1.41] \\
haiku + sonnet & 1.12 & 1.28 (43) & [1.13, 1.43] & 1.09 & 1.30 (52) & [1.16, 1.46] & [1.23, 1.38] \\
opus + sonnet & 1.12 & 1.21 (49) & [1.08, 1.36] & 1.15 & 1.22 (53) & [1.09, 1.36] & [1.16, 1.27] \\
gpt55 + haiku (cross) & 1.31 & 1.40 (6) & [1.16, 1.70] & 1.07 & 1.34 (31) & [1.18, 1.52] & [1.25, 1.44] \\
gpt55 + opus (cross) & 1.31 & 1.40 (6) & [1.16, 1.70] & 1.02 & 1.34 (27) & [1.18, 1.52] & [1.25, 1.45] \\
gpt55 + sonnet (cross) & 1.41 & 1.57 (5)$\dagger$ & [1.29, 1.94] & 1.17 & 1.44 (32) & [1.27, 1.63] & [1.35, 1.55] \\
three same-vendor judges & 1.08 & 1.47 (26) & [1.29, 1.67] & 1.13 & 1.46 (32) & [1.29, 1.65] & [1.36, 1.59] \\
four judges & 1.61 & 1.79 (5)$\dagger$ & [1.47, 2.22] & 1.13 & 1.65 (19) & [1.44, 1.90] & [1.50, 1.84] \\
\midrule
rules + gpt55 & 1.36 & 1.80 (2)$\dagger$ & [1.39, 2.47] & 1.08 & 1.88 (10) & [1.60, 2.22] & [1.67, 2.20] \\
rules + haiku & 1.06 & 1.94 (9) & [1.64, 2.30] & 1.01 & 1.86 (14) & [1.60, 2.16] & [1.67, 2.14] \\
rules + opus & 1.06 & 2.06 (7) & [1.72, 2.48] & 1.06 & 2.09 (7) & [1.75, 2.53] & [1.83, 2.53] \\
rules + sonnet & 1.06 & 2.13 (10) & [1.81, 2.51] & 1.09 & 2.03 (14) & [1.75, 2.36] & [1.82, 2.34] \\
rules + four judges & 1.69 & 2.25 (2)$\dagger$ & [1.74, 3.09] & 1.14 & 2.77 (2)$\dagger$ & [2.14, 3.81] & [2.29, 3.24] \\
\bottomrule
\end{tabular}
\end{table*}

Vendor diversity among the judges recovers little. The three pairs with gpt55 read
above the same-vendor pairs under PRIMARY (1.40, 1.40 and 1.57 against 1.21 to
1.32), and the gap halves under STRICT (1.34, 1.34 and 1.44 against 1.22 to 1.32). That column's collection, in batches of 90 with no provenance, and its review habit (Section~\ref{sec:verdict-alphabet}) confound the excess. The DeepSeek judge of Section~\ref{sec:cross-vendor-arm}, with
clean provenance, gives four pairs at 1.31 to 1.43 (PRIMARY) and 1.26 to 1.34
(STRICT), inside or within 0.12 of the same-vendor range. We therefore report no cross-vendor advantage on pairs with clean provenance.

The right panel of Figure~\ref{fig:main} shows $\nmult$ by corpus with intervals
and floors. On the pooled data the point estimates separate under both definitions,
and whether the intervals separate depends on their type. Under STRICT the case-level
bootstrap intervals separate, with 0 of the 24 rule-pair against judge-pair
combinations overlapping, and the exact intervals of Table~\ref{tab:nmult} overlap in
2 of 24. Under PRIMARY the counts are 5 and 10 of 24. On each corpus the rule pairs read above the judge pairs and the intervals overlap. On the independent corpus the rule pairs read 1.66 to 1.82 against judge pairs at 1.24 to 1.46, and 23 of the 24 rule-pair against judge-pair interval combinations overlap. Under STRICT, with the rule set of the frozen release that preceded the patches (Section~\ref{sec:setup}), the rule pairs on this corpus read 1.67 to 1.74; the difference comes from one block case whose verdict the patches changed. On the external corpus the rule pairs read 1.75 to 2.05 against 1.26 to 1.62, with 24 of 24 overlapping. On the internal corpus every rule-plus-judge stack is censored (0 of 129 joint misses), and the censored lower bounds of every rule pair, 1.39 to 1.56, lie above every judge-pair point
estimate, 1.17 to 1.31, while 22 of 24 interval combinations
overlap. The corpora
taken one at a time are consistent with the split and cannot establish it on their
own.

\subsection{Decoupling Is Not Usefulness}\label{sec:decoupling-not-usefulness}
$\nmult$ is normalised by each layer's own marginal miss rate, so it measures
independence up to the floor and says nothing about competence. On the held-out
corpus the rule layer misses 52.9\% of block cases (46 of 87), against 6.2\% and 6.1\% on the other two corpora. It is the cloud, identity and CI corpus, which the default rule set does not cover.
The rule layer's decoupling is measured on the pooled data, where its pairs sit
0.86 to 1.10 of the way from floor to independence. On the external corpus alone
the reading is weaker. The rule pairs read 1.75 to 2.05 against floors of 1.54 to
1.73, and the judge pairs 1.26 to 1.62 against floors of 1.07 to 1.31. The distances above floor overlap, and 87 cases do not separate the groups. The external corpus shows the competence side without qualification. A stack's
leak rate is the product of both factors, and a layer that misses 53\% of block
cases, however independent, still lets through about half of what its partner misses. This paper should not be read as saying that adding a
regular-expression layer is sufficient.

The competence gap also explains a result we withdrew. Rule-plus-judge stacks beat every judge-plus-judge stack on the independent corpus (12 of 12 paired comparisons significant under STRICT). On the held-out corpus the rule stack is favoured in 1 of 12 comparisons under STRICT and 0 of 12 under PRIMARY, and 9 of 12 and 4 of 12 comparisons reverse direction between corpora. The paired comparison measures independence times competence, so it cannot certify independence on its own.

\subsection{Where the Coupling Comes From}\label{sec:where-coupling-comes-from}
Two candidate factors were separated by construction. Common difficulty is measured
per case as the fraction of a probe set of other layers that err on it, leaving both
members of the pair out, so the score carries no information about the pair's joint outcome. This is the stratifier the prior measurement uses \cite{alotaibi2026layered}. A shared surface mechanism is measured from the case
text alone: an obfuscation device flagged with no reference to any layer's
behaviour. The concealment flag covers encoded and indirect forms of a command (65
of 560 block cases); the broader flag adds eval or exec wrappers and remote
execution (126 of 560).

The mechanism test runs in the direction opposite to the hypothesis. Stratifying by
difficulty and testing whether a device raises joint failure
(Cochran--Mantel--Haenszel) gives median common odds ratios of 0.00 for the
concealment flag and 0.17 for the broader flag across the six judge pairs. Of the
twelve tests, eleven point below 1 and two are significant, both below
1. Cases that carry a device defeat two judges at once less often than cases that do not. The concealment-flag test rests on 0 to 3 joint misses in its exposed stratum, so it can rule the hypothesis out only at
the strengths those counts allow. The prior measurement, with the same stratifier,
recorded the mechanism question as not identifiable, and our data support the same
wording. Difficulty alone can produce coincident failures between independently
built versions \cite{eckhardt1985theoretical,littlewood1989conceptual}. The rule
layer's pairs are excluded from this test. The device flags are lexical and the
rule layer is a lexical matcher, so those pairs are predisposed toward an effect by
construction.

The difficulty share is not identifiable either. We estimate it twice. The first
implementation, in the released analysis script, builds the difficulty score from
the layers outside the pair in a fixed order that admits the archived faithful
DeepSeek column, a column excluded from every other headline number. An earlier draft read its output as a lower bound, and that reading is withdrawn
(Appendix~\ref{app:tables}). The second implementation uses primary layers only.
With the other two judges as the probe, the share of the excess joint failure
attributed to difficulty is 53.9\% (jackknife 95\% interval 45.0 to 62.7) under
PRIMARY and 61.8\% (54.9 to 68.7) under STRICT. With the rule layer added to the probe it is 31.8\% (21.3 to 42.3) and 48.1\% (39.7 to
56.6). The two implementations agree
where their constructions coincide (31.8\% against 32\%). The share moves with the composition of the probe, because the
rule layer's misses carry little information about which cases are hard for judges
and dilute the score. Nothing in these data fixes the share, and the residual
cannot be read as evidence that a shared mechanism exists.

\subsection{What Could Not Be Established}\label{sec:not-established}
Equivalence testing was pre-registered to upgrade ``no coupling detected'' to
``independence established''. The pre-registration is not blind. It was written
after the main analysis had been run on the archived panel, so that panel's four
rule-pair odds ratios were known, and before any judge column was re-collected. What
it fixes is weaker but checkable: bounds derived only from anchors outside this
study, recorded before any equivalence statistic was computed. One known odds ratio,
2.19 for rules with gpt55, lay outside the main bound, and the bound was kept. Both
bounds failed, and we report that instead of substituting a bound that passes.

The main bound, an odds ratio within a factor of 2, is not established for any rule
pair. For three of the four pairs the exact 90\% interval is wider than the
equivalence region even when the odds ratio is exactly 1. The fourth, rules with sonnet, is powered, and its interval $[0.36, 1.35]$ crosses the lower boundary
0.50. Under STRICT its interval, $[0.51, 1.59]$, lies inside the region; the test passes before correction ($p = 0.046$) and fails after Holm correction over the four pairs ($p = 0.184$). The secondary bound, $|\varphi| < 0.30$,
anchored on the smallest $\varphi$ the prior measurement reports, and its negative
control, the six judge pairs run through the same test, were both
pre-registered. All four rule pairs pass the secondary
bound. So does a demonstrably coupled judge pair (gpt55 with sonnet, odds ratio
8.5), and the pre-registration states that a bound such a pair clears cannot
certify independence. The control
disqualified the bound, as the pre-registration provided.

What the data support is the interval. The rule pairs' odds-ratio 90\% intervals
are $[0.31, 10.02]$, $[0.56, 2.32]$, $[0.37, 1.79]$ and
$[0.36, 1.35]$. The correct statement is that no
coupling was detected and the point estimates of $\varphi$ sit near zero. Independence was not
established.

\subsection{An Accounting Rule in the Verdict Alphabet, Not a Security Property}\label{sec:verdict-alphabet}
The verdict alphabet contains one option, review, that is neither an execution nor
a stop. Layers differ in how readily they reach for it. In our panel one judge
returns review on 47 of 560 block cases; the rule layer returns it on 3 and the
other three judges on 2, 12 and 17. That is a
difference in output vocabulary. It interacts with the accounting rule for review and reflects no property of the
model. We name the layers elsewhere for
reproducibility; the finding below is about the accounting rule, and reading it as a
vendor comparison would be a misreading.

Any metric that credits escalation as a catch scores a deferring layer well. In a
stack with no human layer to absorb the escalation, that layer contributes nothing.
In this study the choice of accounting rule reversed a substantive conclusion five
times, each time in the direction that made the deferring layer look better
(Table~\ref{tab:reversals}).

\begin{table}[t]
\caption{Five conclusions that reverse between the two accounting rules for review
(PRIMARY: review counts as caught; STRICT: review counts as not stopped).}
\label{tab:reversals}
\centering
\footnotesize
\begin{tabular}{@{}p{1.8cm}p{3.5cm}p{2.6cm}@{}}
\toprule
where & PRIMARY $\rightarrow$ STRICT & effect \\
\midrule
the deferring layer's solo miss & 0.014 $\rightarrow$ 0.098 & the most accurate judge becomes an ordinary one \\
its pairwise correlation & $\varphi$ 0.146 to 0.253 $\rightarrow$ at least 0.328 & the weakest-coupled layer joins the rest \\
paired stack comparison & 6 of 12 significant $\rightarrow$ 12 of 12 (0 against under either) & a partial result becomes complete \\
value of a fourth judge & 21 cases added $\rightarrow$ 13 & 8 of the 21 were deferrals \\
$\nmult$, pairs with the deferring judge & 1.40 / 1.40 / 1.57 $\rightarrow$ 1.34 / 1.34 / 1.44, against same-vendor pairs at 1.21 to 1.32 & the apparent cross-vendor gap halves \\
\bottomrule
\end{tabular}
\end{table}

The fourth-judge instance is the cleanest. Of the 21 cases the deferring judge appeared to
add to the three same-vendor judges, 13 were block verdicts and 8 were
review. The rule layer's advantage over a fourth judge in Table~\ref{tab:increments} reversed on the archived panel and does not reverse on this panel (Section~\ref{sec:archived-panel}). Whether a convention flips a conclusion depends on the panel it is applied to.

An evaluation convention rewarded behaviour that added nothing to the stack, and
only the stack-level measurement showed it. The correct accounting for review
depends on the human layer we do not have (Section~\ref{sec:limitations-m4}, item
6).

%% file: sections/06_replication.tex
\section{Comparison Panels and Provenance}\label{sec:replication}

The primary panel of Section~\ref{sec:setup} was re-collected for this paper. We compare it with five other collections of the same cases under the same case-text condition, report what the collections revealed about provenance, and then run the layers on a corpus from another team. The correlation structure and the separation of the $\nmult$ bands reproduce in the archived panel, in a version-pinned re-collection and for the third-vendor judge. For the memory-index panel and the single-case check we report verdict agreement, which measures stability rather than structure. The marginal miss rates do not reproduce, and they were not
expected to.

\subsection{The archived panel}\label{sec:archived-panel}
Three archived columns for the same three tiers carried an earlier draft. The operator recorded them as Claude Haiku 4.5, Claude Sonnet 4.6 and Claude Opus 4.8 and collected them in batches of 90 in identifier order with no version, timestamp or temperature recorded. They agree with the re-collected columns on 0.814, 0.860 and 0.895 of verdicts
(haiku and sonnet on all 1{,}119 cases; opus on the 620 cases whose batches were
served by the requested version). The comparison holds the
case blocks fixed and lets the instruction header differ, as the DeepSeek drift comparison below does. The archived headers carried a list of example threat
patterns that the re-collection header drops (Section~\ref{sec:setup}).
Their structure is the same: rule pairs' $\varphi$ median $+0.008$ (PRIMARY) and
$+0.011$ (STRICT), judge pairs' $+0.420$ and $+0.540$, rule-pair $\nmult$ of 1.80
to 2.13 and judge-pair $\nmult$ of 1.19 to 1.52 across the two
definitions. Their marginals are not the same. The
archived sonnet misses 0.068 of block cases against 0.170 for the re-collected
sonnet tier (PRIMARY, discordant 15 and 72, McNemar $p < 0.001$). Haiku (0.104 against 0.105) and opus (0.108 against 0.086 on the 279 same-version block
rows) are indistinguishable. One conclusion did depend on the panel. On the archived panel the
rule layer's advantage over a fourth judge in Table~\ref{tab:increments} was a STRICT-only result (68 against 21 under STRICT, 20 against 22 under PRIMARY). On the primary panel it holds under both definitions (86 against 22, 31 against 21).

\subsection{A Version-Pinned Re-collection}\label{sec:pinned}
The three tiers were collected a third time with the version pinned on every call,
to the versions the operator recorded for the archived panel. The served model
equalled the requested \code{claude-opus-4-8}, \code{claude-sonnet-4-6} and
\code{claude-haiku-4-5-20251001} on every call, and all 1{,}119 cases were covered;
six single-case judgments of the haiku tier are missing. Each tier ran in its own
session, and the tasks were delivered as files rather than inline, unlike the primary
panel. The rule and gpt55 columns are those of the primary panel.

\begin{table}[t]
\caption{Marginals of the version-pinned re-collection against the primary and the
archived panels ($n = 560$ block cases for misses, 379 allow cases for false
alarms). PRIMARY counts review as caught; STRICT counts it as not stopped. The
primary panel's sonnet tier was served as \code{claude-sonnet-5} and its opus tier
partly as \code{claude-opus-4-8} (Section~\ref{sec:version-changes}); the archived
versions are as recorded by the operator.}
\label{tab:pinned}
\centering
\footnotesize
\setlength{\tabcolsep}{3pt}
\begin{tabular}{@{}lrrrrrrr@{}}
\toprule
 & \multicolumn{3}{c}{pinned} & \multicolumn{2}{c}{primary} & \multicolumn{2}{c}{archived} \\
tier & miss P & miss S & FA & miss P & FA & miss P & FA \\
\midrule
opus & 72 & 72 & 18 & 59 & 16 & 48 & 4 \\
sonnet & 21 & 31 & 76 & 95 & 4 & 38 & 6 \\
haiku & 49 & 65 & 13 & 59 & 18 & 58 & 9 \\
\bottomrule
\end{tabular}
\end{table}

Two findings follow from Table~\ref{tab:pinned}. The primary panel's high sonnet
miss count comes mostly with the version: its \code{claude-sonnet-5} misses 95 block
cases under PRIMARY, against 21 for the pinned \code{claude-sonnet-4-6} and 38 for the
archived column recorded as the same version.
Under the same recorded version, a change of collection conditions moves single-layer
numbers: opus misses 72 block cases against 48 on the archived panel, and sonnet
false-alarms on 76 of the 379 allow cases against 6. The pinned tiers agree with the
archived verdicts on 1{,}003, 892 and 938 of 1{,}119 cases (opus, sonnet, haiku) and
with the primary panel on 1{,}043, 828 and 950.

The structure does not move. Under STRICT the rule pairs have $\varphi$ of $-0.047$ to
$+0.041$ and compose to 1.88 to 2.18 layers above floors of 1.02 to 1.19, and the
judge pairs have $\varphi$ of 0.34 to 0.58 and compose to 1.20 to 1.37 layers above
floors of 1.02 to 1.17. On the normalised scale the rule pairs read 0.87 to 1.18 and
the judge pairs 0.03 to 0.35, against 0.86 to 1.10 and 0.08 to 0.33 on the primary
panel. Neither the exact nor the bootstrap intervals of the two groups overlap (0 of
24 combinations each). Under PRIMARY the rule pairs have $\varphi$ of $-0.05$ to
$+0.041$ and read 1.80 to 2.39 layers above floors of 1.01 to 1.36, the judge pairs
0.06 to 0.54 and 1.24 to 1.69 above floors of 1.09 to 1.35. Under PRIMARY 15 exact
and 7 bootstrap interval combinations of 24 overlap, all in pairs with gpt55. Three pairs
have only one or two joint misses under PRIMARY (rules with sonnet, rules with gpt55,
gpt55 with sonnet), so their normalised readings are unstable. The same recorded
versions under different collection conditions give different marginal rates and the
same structure.

\subsection{Blinding at the system level}\label{sec:memory-panel}
A first re-collection was run inside the project workspace with the same prompts. A
probe of the subagent system context found that the harness injects a fixed project-memory index of nine entry titles. These name the runtime, the target venue and the two elicitation conditions but contain no hypothesis, no other judge's
verdict and no case label. The primary panel was therefore collected again in a
session without that memory directory, with prompts, seed, batch split and
single-case sample identical to the first re-collection by hash. A probe of its system context found only two unrelated entry titles and a generic harness skill
description. On
the external corpus the two collections agree case for case on 0.847, 0.921 and
0.947 of four-way verdicts (haiku, sonnet, opus), on 0.931, 0.974 and 0.984 of
PRIMARY miss labels and on 0.974, 0.968 and 0.979 of STRICT miss
labels. The injected
index moved a few verdicts and no headline. The prompt-level blindness assertion
(17 banned terms, 0 hits) held for both collections. An earlier claim that the judges had seen no mention of this study was withdrawn when the probe showed the
index (Appendix~\ref{app:tables}).

\subsection{Single-case judging against batched judging}\label{sec:batch-vs-single}
Each primary tier judged 100 sampled cases a second time, one case per call, with
the same instruction. Exact four-way agreement with the batched verdict was 0.840,
0.837 and 0.900 (haiku, sonnet, opus), agreement on the PRIMARY miss label 0.880, 0.888 and 0.940, and on the STRICT miss label 0.910, 0.898 and 0.940. Sonnet has 98 cases, because two calls failed twice by echoing the instruction's example identifier. Batching changes about one verdict in ten and about one miss label in ten or
fewer.

\subsection{Served-Model Changes and a Pre-Declared Rule Set Aside}\label{sec:version-changes}
The collection pre-declared that a batch served by a model other than its tier's
majority version would be marked mixed-version, that the majority version would
carry the primary analysis, and that all batches would be reported as a sensitivity
configuration. The rule assumed version changes would be rare, and they were not. Of 112 opus batches, 50 were completed by \code{claude-opus-4-8} while the requested
model was \code{claude-opus-5}, and the majority-version rule would have left 279
block rows with all five layers. We therefore report the full-batch panel as primary and the
majority-version subset as the sensitivity configuration, and we split the opus
tier by served version. Of the 50 batches, 12 followed a refusal event in the
transcript, 2 show a version sequence without a refusal, and 36 show no signal of
any kind; the change is visible only in the per-call served-model
field. The haiku and sonnet tiers were served by
one version on all 112 batches.

The change was not uniform across the corpora. Counting block rows whose opus
verdict came from \code{claude-opus-4-8}: 69 of 87 on the external corpus (79\%),
171 of 344 on the independent corpus (50\%) and 41 of 129 on the internal corpus
(32\%), $\chi^2(2) = 47.0$, $p = 6.1 \times 10^{-11}$ (over all 1{,}119 cases:
129 of 189, 280 of 630 and 90 of 300, $\chi^2(2) = 68.7$, $p = 1.2 \times
10^{-15}$). In this collection the served version was strongly associated with the
corpus, which is the question that two events in an earlier re-collection could not answer. We report the counts and the test and infer nothing about the routing mechanism. Both served-version subsets keep the rule layer decoupled from the opus
tier: rules $\times$ opus $\varphi = +0.047$ on the 279 \code{claude-opus-5} block
rows and $-0.069$ on the 281 \code{claude-opus-4-8} rows, with $\nmult$ of 1.83 and
2.26 under PRIMARY and 1.85 and 2.31 under
STRICT. One batch made one harmless tool call and three drew a harness classifier warning
with no effect on the output; all are kept.
The archived gpt55 column has no served-model field, so events of this kind there
are invisible.

\subsection{Cross-Vendor Re-collection with Recorded Provenance}\label{sec:cross-vendor-arm}
The DeepSeek judge was re-collected over the external (189) and independent (630)
corpora, one case per request, temperature 0, 200 output tokens, JSON-object
response format, under two conditions: no-tell, with case blocks byte-identical to
the archived no-tell files, and faithful, with the human-written description shown.
There were 1{,}638 requests and 1{,}638 HTTP 200 responses, with no retry
recorded. Every record carries the
served model, a UTC timestamp, the prompt hash and the raw response. The requested
alias was \code{deepseek-chat}; the served model was \code{deepseek-v4-flash} on all
1{,}638 calls. An alias in a configuration file is not a version.

The cross-vendor judge lands in the judge band. On the 431 block cases of the two
re-collected corpora the no-tell column couples with all four primary judges and
does not couple with the rule layer (Table~\ref{tab:recollect}). With the DeepSeek
column included, the rule pairs' $\varphi/\varphi_{\max}$ run $-0.045$ to $+0.154$
(0 of 5 significant) and the judge pairs' $+0.390$ to $+0.847$ (10 of 10) under
PRIMARY; under STRICT the ranges are $-0.024$ to $+0.087$ against $+0.395$ to
$+0.824$ (10 of 10). The bands
stay separated.

\begin{table*}[t]
\caption{The re-collected DeepSeek column against the five primary layers
(external and independent corpora, $n = 431$ block cases). Every judge pair here is
cross-vendor. Significance is Fisher's exact test with Holm correction over the
fifteen pairs of the six-layer panel, given for PRIMARY / STRICT.}
\label{tab:recollect}
\centering
\small
\begin{tabular}{@{}lrrrrrrl@{}}
\toprule
 & \multicolumn{3}{c}{PRIMARY} & \multicolumn{3}{c}{STRICT} & \\
pair & $\varphi$ & $\varphi/\varphi_{\max}$ & $\nmult$ & $\varphi$ & $\varphi/\varphi_{\max}$ & $\nmult$ & sig. \\
\midrule
rules + DeepSeek & $+0.057$ & $+0.076$ & 1.84 & $+0.062$ & $+0.080$ & 1.83 & no / no \\
gpt55 + DeepSeek & $+0.271$ & $+0.684$ & 1.38 & $+0.501$ & $+0.508$ & 1.26 & yes / yes \\
haiku + DeepSeek & $+0.388$ & $+0.424$ & 1.35 & $+0.440$ & $+0.525$ & 1.32 & yes / yes \\
opus + DeepSeek & $+0.438$ & $+0.479$ & 1.31 & $+0.404$ & $+0.434$ & 1.34 & yes / yes \\
sonnet + DeepSeek & $+0.333$ & $+0.491$ & 1.43 & $+0.438$ & $+0.657$ & 1.34 & yes / yes \\
\bottomrule
\end{tabular}
\end{table*}

Every judge pair in Table~\ref{tab:recollect} is cross-vendor, and every one
couples at the strength of the same-vendor pairs of Section~\ref{sec:readings}
\cite{goel2025alike,kim2025correlated}. The pair of the OpenAI judge with DeepSeek,
two vendors and two independent collection chains, gives $\nmult = 1.26$ under
STRICT, with a 95\% interval of $[1.09, 1.45]$. The rule pair gives 1.83 with
$[1.54, 2.18]$. Vendor diversity
does not recover independence. Per corpus the rule pair is $+0.134$ and $+0.111$
(PRIMARY) and $+0.097$ and $+0.107$ (STRICT), the largest rule pairs in the study,
none significant.

\paragraph{Drift and elicitation}
Against the archived DeepSeek column (faithful condition, version unrecorded) the
re-collected faithful column agrees on 692 of 819 verdicts (0.845). Block miss
rates are indistinguishable, 0.077 against 0.081 under PRIMARY (discordant 13 and
15, McNemar $p = 0.85$ \cite{mcnemar1947note}) and 0.081 against 0.081 under STRICT. The false-alarm rate fell from 0.085 to 0.033 (16 and 3, $p = 0.004$),
almost entirely on the external corpus (0.310 to 0.086, 13 and
0). The comparison is condition-matched and
not prompt-identical, so it bounds drift plus wording from above: miss behaviour is stable across a model version, and false-alarm behaviour is not. Under the same served
model in the same hour, description blanked against description shown agrees on
718 of 819 verdicts (0.877). On the independent corpus the description changes
nothing (miss 0.105 against 0.102 under PRIMARY, $p = 1.0$, and false alarms 0.021 against 0.016). On the external corpus it removes every miss (5 to 0 under PRIMARY
with $p = 0.062$, 7 to 0 under STRICT with $p = 0.016$) and most false alarms (0.224
to 0.086, 8 and 0, $p = 0.008$). On that
corpus the human-written description is a tell that carries enough of the label to zero one judge's miss rate. That quantifies why the primary layer set is measured no-tell (Section~\ref{sec:setup}) and why that field must not be shown to
any layer whose independence from the labels is being measured.

\subsection{A Corpus from Another Team}\label{sec:redcode}
All three corpora come from the runtime's project. As a bound on how far the reading
travels, the rule layer and the three tiers at the pinned versions were also run under
the same no-tell elicitation on the bash subset of RedCode-Exec
\cite{guo2024redcode}: 598 risky bash code snippets after two byte-identical
duplicates were removed, all scored against a block label. Its opus and sonnet
versions differ from the primary panel's. For four haiku batches (40 cases) a platform
filter stopped every response, so the missingness is not random, and the statistics
below use the 558 cases that all three tiers judged. Every layer misses most cases:
under STRICT the rule layer misses 431 of them and the judges 336 to 407. On the
normalised scale the rule pairs read 0.43 to 0.49 and the judge pairs 0.11 to 0.19
under STRICT. The order of the two groups holds, but the rule pairs sit only midway
between full coupling and independence. After stratification by difficulty
(Section~\ref{sec:where-coupling-comes-from}), two of the three rule pairs keep no
detected residual association, rules with opus keeps a significant one, and the judge
pairs stay coupled; under PRIMARY, difficulty accounts for 0.76 to 0.91 of the rule
pairs' excess joint failure against 0.58 to 0.69 for the judge pairs.

%% file: sections/07_related_work.tex
\section{Related Work}\label{sec:related}

\paragraph{Correlated failure in defence stacks}
The closest prior measurement is a seven-layer single-turn jailbreak-defence stack
\cite{alotaibi2026layered}: two token-surface filters, a safety classifier, a
refusal-priming prefix, a perturbation-stability check and two hidden-state probes,
all on one open-weight target model under one adaptive adversary. It reports
positive $\varphi$ in all fifteen measurable pairs (0.30 to 0.75), a joint residual
above the multiplicative prediction, and a difficulty stratification under which
almost every cross-row association dissolves. Across three labelling regimes the
surviving pair changes, so the authors retire the mechanism-specific reading and
record the question as not identifiable.  Its scope is described as single-turn jailbreak only,
input-provenance and architecture rows are excluded because they target injection,
every measured member is a function of the target model, and correlation across
agentic settings is listed as unmeasured. Only one of its seven members is an LLM
judge, and none is a hand-authored rule layer. Our judge pairs reproduce its finding
in a new setting and carry no novelty claim. Our difficulty decomposition uses the
same stratifier and reaches the same non-identifiability. In our judgement our rule
pairs sit at a larger mechanism distance than any pair it measured, though that
study defines no such quantity and the comparison is ours. Its correlations were
collected under an adaptive adversary and ours were not
(Section~\ref{sec:threat-model}).

Correlated errors across LLMs from different vendors have been measured before. Models that share more errors are found to be poorer overseers of one another
\cite{goel2025alike}, and models from different vendors agree on a majority of the
cases on which both err \cite{kim2025correlated}. Our cross-vendor arm
(Section~\ref{sec:cross-vendor-arm}) reproduces that finding for judges of agent
actions. Common-cause failure in reliability engineering is treated with ratio
parameters, the beta factor and its refinements \cite{iec61508}. Eckhardt and Lee
showed that variation in per-case difficulty is by itself sufficient to make
independently developed versions fail together \cite{eckhardt1985theoretical}.
Knight and Leveson measured the failure of the independence assumption in
multi-version programming \cite{knight1986experimental}. Littlewood and Miller
modelled coincident failures under diverse development
\cite{littlewood1989conceptual}, and Littlewood, Popov and Strigini review the
design-diversity models that followed \cite{littlewood2001diversity}. The same
question has been measured for security detectors: Bishop et al.\ and Gashi et al.\
measured how much diverse off-the-shelf antivirus engines gain from being combined,
on the same inputs \cite{bishop2011diversity,gashi2013malware}. That is our
question, asked of an earlier class of detector. Classifier-ensemble research relates diversity
measures to ensemble accuracy \cite{kuncheva2003diversity}. Hill numbers supply the
effective-count construction we reuse
\cite{macarthur1965patterns,hill1973diversity,jost2006entropy,jost2007partitioning}.
Kish's effective sample size \cite{kish1965survey} has been applied to LLM judge
panels \cite{kohli2026ninejudges} and is a different quantity
(Section~\ref{sec:instrument}).

\paragraph{Relation to the prior reports on these data}
The runtime, the rule layer and the corpora come from two technical reports
\cite{agenttrust2026a,agenttrust2026b}. The second reports, on the same external
corpus, the rule layer's 52.9\% miss rate, the cloud pack's zero change on the
semantic categories, and the clustering of three judges' accuracies between 83.6\% and 85.2\%. The first reports the joint miss rate of a rules-plus-judge hybrid on
its benchmarks. This paper adds the pairwise correlation structure,
$\nmult$ with its floor, the difficulty decomposition, the two evaluation
conventions, the cross-vendor column, and the case-level accounting of what each
increment adds to a stack.

\paragraph{Enforcement theory}
Ray characterises which policies a deterministic gate can enforce relative to an
oracle interface, gives decidability and complexity results for guard theories, and bounds the robustness cost of attacks on a judge's visible representation \cite{ray2026enforced}. The bound is stated as distributional, with no claim of universal semantic soundness. It asks what is enforceable
given an interface; we measure how the errors of several concrete layers relate.

\paragraph{Enforcement architectures}
Several designs keep injected content away from the action space: CaMeL through
restricted code with capability checks before each tool call
\cite{debenedetti2025camel}, FIDES through taint tracking with deterministic policy
enforcement \cite{costa2025fides}, Progent through symbolic privileges over tool
names and arguments \cite{shi2025progent}, and Parallax through tiered verification
behind a deliberation-action split, which also names the assume-compromise
evaluation we adopt \cite{fokou2026parallax}. Guardrail systems classify prompts,
outputs or agent traces
\cite{inan2023llamaguard,chennabasappa2025llamafirewall,rebedea2023nemo}, and the
AgentTrust runtime intercepts actions with a rule gate, an opt-in judge and a human
tail \cite{agenttrust2026a,agenttrust2026b}. These works place enforcement points.
We did not find, in any of them, a measurement of how the errors of one point relate to the errors of another. This paper adds that measurement.

\paragraph{Evaluation of judges and agents}
Benchmarks measure how often an agent or a judge is fooled: AgentDojo for prompt
injection in tool-using agents \cite{debenedetti2024agentdojo}, AgentHarm for
harmful agent tasks \cite{andriushchenko2024agentharm}, ToolEmu for emulated tool
risk \cite{ruan2024toolemu}, R-Judge for judging agent risk \cite{yuan2024rjudge},
and AgentTrap for runtime trust failures in third-party skills
\cite{zhuang2026agenttrap}. Indirect prompt injection itself was characterised by
Greshake et al.\ \cite{greshake2023injection}, and adaptive attacks against
published defences and judges are the subject of \cite{nasr2025attacker} and
\cite{zhan2025adaptive}. RobustJudge reports that no evaluated defence dominates on
robustness, benign utility and cost at once, and that judge robustness is highly
sensitive to prompt template and model choice \cite{li2025robustjudge}. That
sensitivity bounds how far any single-template result of ours generalises, and we
cite it for that reason. These works measure whether a judge is fooled; we measure
whether different layers are fooled by the same cases. The human layer we lack has
its own literature, in which oversight capacity and fatigue have been modelled
\cite{turan2026oversight}.

%% file: sections/08_discussion.tex
\section{Discussion}\label{sec:discussion}

\paragraph{What a practitioner should do with the measurement}
Select stack members on mechanism class, then measure the assembled stack. The
pairwise statistics $\varphi$ and $\nmult$ are selection statistics. They say which
candidate layers are worth composing, and they do not predict the assembled
stack's residual, which we measured directly as added coverage. This two-stage
workflow is the one the prior measurement recommends \cite{alotaibi2026layered},
and our data agree with it. Two corollaries follow from
Section~\ref{sec:instrument-fails}. A layer's solo accuracy is the wrong number to
report when the question is what the layer adds. Improving a layer along its own
mechanism axis can leave the stack unchanged and move the layer toward the coupled
band, because the gains land on cases the other layers already catch.

\paragraph{Three claims that must stay separate}
The result invites a transfer from redundancy engineering, and the transfer has
three parts that carry different weight. We measured the statistical part: on these corpora, without an adaptive adversary, errors across mechanism classes showed no correlation we could detect, and errors
within the judge class were correlated. The complexity part,
that a deterministic layer is cheap to run and audit and a judge is not, is a
property of the implementations and is stated here without measurement. The
expressiveness part, that a regular-expression layer cannot recognise the threats a
judge recognises, is the reason the rule layer misses 52.9\% of the external corpus. It is an argument about what each mechanism can see, and this study does not test it.

\paragraph{Two conventions as requirements}
The accounting rule for review verdicts reversed five conclusions during this
study, and the served-model log overturned a pre-declared analysis rule. Both are cheap to adopt. A stack
evaluation should state which verdicts count as a stop, report every headline under
both readings when a deferring layer is present, and record the served model
string on every call. We make no claim about why a served model changes mid-collection. In this study
one tier was served by another version in 50 of 112 batches, 36 of them with no
signal in the transcript, and the change fell on 79\% of the external corpus's block rows against 32\% of the internal corpus's. Per-call logging shows that a version changed and which content the change fell on, the subset a safety evaluation cares about. The pre-declared rule that the majority version
would carry the analysis is a convention the data overruled, and the deviation is
disclosed (Section~\ref{sec:version-changes}). Which accounting rule applies to a
deployment depends on whether a human layer absorbs escalations. Where one exists, the PRIMARY reading is the relevant one, and the gap between the archived judge's
pairs and the same-vendor pairs is then at its widest, though it is confounded with
that column's collection (Section~\ref{sec:readings}).

\paragraph{Mechanism distance}
Mechanism distance is sampled at two points, regular-expression matchers against
chat models, and the weaker deterministic layer cannot be read on the independence
axis (Section~\ref{sec:instrument-fails}). We name the split a boundary
observation. A gradient would need intermediate mechanisms, such as a typed
capability gate or a fine-tuned classifier, measured with the same instrument.

\paragraph{Scope of the corpora and the judges}
The independent corpus's categories are verdict tiers, and it was not built to
contain semantic threats. The results that concern semantic threats all come from the external corpus, and
a reader should not take the 630-case corpus as evidence about them. Every judge here ran one prompt template, and judge robustness depends on the template \cite{li2025robustjudge}. A different template could move the marginal rates, though the correlation structure appears in the archived panel, in the
version-pinned re-collection and for the third-vendor judge
(Section~\ref{sec:replication}).

%% file: sections/09_limitations.tex
\section{Limitations}\label{sec:limitations-m4}

\begin{enumerate}
\item Mechanism distance is sampled at one axis with two points, and the second
  deterministic layer cannot be read on that axis. The attribution of decoupling to
  the mechanism class rests on the default rule layer across three corpora, the
  comparison panels and the cross-vendor column; b1's pairs sit on their floors
  (Section~\ref{sec:instrument-fails}). Deterministic mechanisms that are not
  regular expressions, such as type systems and capability gates, are untested.
  Two configurations do not make a gradient, and none is claimed.
\item Full-stack $\nmult$ is censored. Across 560 block cases all five layers miss
  together twice under either definition, and on the internal corpus never.
  Pairwise $\nmult$ is well estimated where $k$ is large; rows with $k \le 5$
  support bounds only, and the two-miss stacks are censored in 13.7\% of bootstrap
  resamples.
\item The difficulty share is not identifiable
  (Section~\ref{sec:where-coupling-comes-from}). It ranges from 31.8\% to 61.8\%
  with the probe used to score difficulty and the miss definition, and the residual
  must not be read as evidence of a shared mechanism. The concealment-flag test
  rests on 0 to 3 joint misses in its exposed stratum.
\item The internal corpus cannot distinguish stack configurations (0 of 129 joint
  leaks for every rule-plus-judge stack), and the interval evidence for the split
  comes from the pooled data only; on each corpus taken alone the intervals overlap
  (Section~\ref{sec:readings}).
\item Four judges, three from one vendor. The cross-vendor evidence has two sources. The archived gpt55 column's pairs read 0.02 to 0.12 above the same-vendor pairs under STRICT, in a way confounded with its collection and its review habit. The DeepSeek arm of Section~\ref{sec:cross-vendor-arm}, collected one case per request with the served model recorded, gives four judge pairs at $\nmult$ of 1.26 to 1.34 under STRICT and 1.31 to 1.43 under PRIMARY, while its rule pair reads 1.83 and 1.84. We report no cross-vendor advantage on the pairs with clean provenance. A judge that is not a general-purpose chat model remains untested.
\item No human review layer. How review should be scored therefore cannot be settled
  here, and both readings are reported throughout (Section~\ref{sec:setup}).
\item The 180 warn cases are excluded from both error domains.
\item The gpt55 column and the archived comparison panel were collected in batches
  of 90 cases in identifier order, with no record of wrapper prompt, temperature,
  model version or date, and they cannot be regenerated. The primary tiers were collected in batches of 10 by one operator through one
  harness, under an instruction header that differs from the archived headers
  (Section~\ref{sec:setup}), so the archived-panel comparison is condition-matched on the case blocks only. Batching changes about one verdict in ten and about one
  miss label in ten or fewer (Section~\ref{sec:batch-vs-single}). The version-pinned
  re-collection ran each tier in its own session with the tasks delivered as files
  (Section~\ref{sec:pinned}), so its comparison with the primary panel is also
  condition-matched on the case blocks only. The runtime's rewrite-type safe-alternative
  component sits outside the measured stack and is untested here.
\item In the primary panel a dated snapshot identifier exists for the haiku tier only.
  The sonnet and opus tiers carry aliases, the opus alias was served by two versions,
  and the gpt55 column has no provenance at all (Section~\ref{sec:replication}). The
  version-pinned re-collection requested and received a fixed version on every call
  (Section~\ref{sec:pinned}).
\item The zero-new-coverage result for the cloud pack is one pack and 17 to 19
  newly caught cases, all but two of them on one corpus. On that corpus it is clean
  (100\% across four judges and both definitions), and it must
  not be generalised to ``rule augmentation always lands on redundant cases.'' The methodological half generalises: solo accuracy did not predict stack contribution, and one counterexample suffices for that.
\item A defect in the rule layer's elicitation, disclosed. The rule layer's matcher
  reads the action description, so its archived verdicts were produced in the
  faithful condition while the judges ran no-tell. Every headline quantity was
  recomputed with the description blanked. Blanking it changes the rule verdict on 2
  of 560 block cases, moves the rule pairs' PRIMARY $\varphi$ by at most 0.002 and
  their $\nmult$ by at most 0.02, leaves the pooled STRICT values unchanged, and
  leaves the new-coverage totals unchanged (86 to
  86). The
  recomputed column is part of the material available on request.
\item Seven candidate conclusions were withdrawn during analysis after failing
  replication, equivalence, definition-sensitivity, probe-composition or blindness
  checks; the pre-registration and a summary of the audit trail are in the artifact
  (Appendix~\ref{app:tables}).
\item The external corpus is selected on label agreement. Of its 224 source cases,
  all blind-relabelled by an independent human annotator who is not an author, 193
  agreed with the LLM rubric label and 31 disagreed (8 allow to warn, 13 warn to
  allow, 5 warn to block, 5 block to warn; no allow-to-block flip), with
  $\kappa = 0.817$ before filtering. The frozen 189 are agreed cases \cite{agenttrust2026b}. Selecting on agreement biases the corpus toward clear-label cases, so the 52.9\% miss rate and the judge coupling there are measured on such cases. A residual shared-prior concern remains for cases that both an LLM and a human over-label.
\item All results are distributional statements about labelled corpora without an
  adaptive adversary (Section~\ref{sec:threat-model}). They bound nothing about an
  attacker who targets the joint miss set.
\item One judge tier was served by a version other than the one requested in 50 of
  112 batches. The pre-declared majority-version rule was set aside because it
  would have left 279 block rows; the full-batch panel is primary, the
  majority-version subset is a sensitivity configuration, and the tier is reported
  split by served version (Section~\ref{sec:version-changes}).
\item The third-party corpus of Section~\ref{sec:redcode} is a single source and has
  only block cases, so it measures misses and not false alarms, and scoring warn as a
  miss is our mapping. Its judge versions differ from the primary panel's, 40 haiku
  cases are missing for a reason that is not random, and its difficulty probe has
  fewer layers than in Section~\ref{sec:where-coupling-comes-from}.
\end{enumerate}

%% file: sections/10_conclusion.tex
\section{Conclusion}\label{sec:conclusion}

We measured whether the errors of stacked action gates multiply, on 1{,}119
labelled actions from three corpora, with one rule layer and four judges under one case-text condition. Three of the judges were re-collected with the served model recorded on every call. Across mechanism classes no coupling was detected and the
point estimates sit within reach of the multiplication rule, but independence was
not established; within one class the errors largely do not multiply. Under STRICT, any two judges compose to about 1.2 to 1.4
multiplication-equivalent layers, and a rule layer with a judge to 1.9 to 2.1; the floors set by the marginals are
1.02 to 1.17 and 1.01 to 1.09. The shape held on the pooled
data, was consistent with each corpus, and reappeared in the archived panel, in a
version-pinned re-collection and for a fifth judge from a third vendor. A panel
collected with a memory index in the judges' context, and single-case judging, agreed
with the primary panel on most verdicts. On a third-party corpus on which every layer
misses most cases, the rule pairs sat only midway between full coupling and
independence before stratification, and most of their coupling tracked shared
difficulty. Solo accuracy did not
predict what a layer added to the stack in either direction. Mechanism class and
in-domain competence must both be read. The practical procedure is to select
members by mechanism class and then measure the assembled stack.

Two evaluation conventions changed conclusions in this study, and we recommend both
as requirements for stack evaluations: report every result under an explicit
accounting rule for review verdicts, and log the served model on every call. In this study that log showed one tier served by another version in 50 of 112 batches, concentrated on one corpus. Four things remain unmeasured. A human review layer would settle how review is scored. A
judge that is not a general-purpose chat model would test whether the judge band
is a property of chat models. A deterministic mechanism other than a regular expression would give mechanism
distance a second point. An adaptive adversary would test what the distributional
independence measured here is worth against an attack aimed at the joint miss set.

%% file: sections/90_open_science_arxiv.tex
\section*{Open Science}
The artifact for this paper is available at \artifacturl. It contains the per-layer verdict matrix for all 1{,}119 cases, for the primary panel and for the archived panel. It contains the per-case and per-call provenance of the re-collected judge columns (model requested, model served, timestamps, prompt and output hashes, stop reasons and refusal flags), for the primary collection and for the comparison collection of Section~\ref{sec:memory-panel}, and the re-collected DeepSeek columns under both elicitation conditions with per-call provenance (served model, UTC timestamp and prompt hash). It also contains the per-layer verdicts and per-call provenance of the version-pinned re-collection (Section~\ref{sec:pinned}) and the per-case verdicts and per-call records of the third-party corpus run (Section~\ref{sec:redcode}); the RedCode case texts are available from their authors \cite{guo2024redcode}. It also contains the analysis scripts that compute the statistics of Sections~\ref{sec:readings} and~\ref{sec:replication} from those files, the pre-registration of the equivalence test, summaries of the verification-coverage audit, the instrument-definition audit and the provenance audit, a summary of the recorded deviations, and the record of the conclusions withdrawn during analysis. The scripts are deterministic and rerun to byte-identical outputs, whose hashes the artifact lists. The case texts, the prompts that embed them, the judges' raw responses, the external corpus and the collection scripts are not part of the public artifact, because the dangerous cases are command and code strings written as attacks (see the Ethical Considerations section). They are available from the author on request. The results that need the case texts, namely the cloud rule pack, the second rule engine and the no-tell audit, cannot be recomputed from the public artifact. The rule files and the internal and independent corpora are those of a public release of the AgentTrust runtime \cite{agenttrust2026a}. The archived gpt55 column and the archived comparison panel cannot be regenerated, because their collection recorded no model version, timestamp or temperature.

%% file: sections/91_llm_usage.tex
\section*{LLM usage considerations}

\emph{Transparency.} LLMs are the object of study. The judge layers measured in Section~\ref{sec:setup} are production LLM judges:
one archived column, three columns re-collected through a coding-assistant harness
with the served model recorded on every call, and one column re-collected through a
vendor API. The comparison panels of Section~\ref{sec:replication} are further
collections of the same kind. Their versions, where known, temperatures and query
volumes are stated in those sections. Where a version is not known we say so per
tier and do not claim that it was recorded. The reproducibility limitation this
creates is one of the paper's findings (Section~\ref{sec:version-changes}). Closed
models changed version during collection, which is why the served-model string was
recorded on every call and is released with the artifact.

\emph{Accountability and correctness.} LLMs were used for editorial purposes in
this manuscript, and all outputs were inspected by the author to ensure accuracy
and originality. LLM coding assistants were also used to draft the analysis
scripts. The released analysis scripts rerun to the numbers reported in Sections~\ref{sec:readings} and~\ref{sec:replication}, and seven candidate
conclusions were withdrawn when the audited outputs did not support them. The author takes responsibility for every claim in the paper.

\emph{Responsibility.} No model was trained. Query volume was kept to one judgment
per case and condition: the archived verdict files hold 4{,}776 judgments in 14
files. The cross-vendor re-collection made 1{,}638 API calls with no retry. The primary panel was collected through a coding-assistant harness in 642 calls
(336 batched, 300 single-case, 6 retries), 26{,}273{,}030 input tokens including
cache reads and 1{,}266{,}080 output tokens, with 12 refusal events (all on one
tier, each completed by a second served version) and one tool
call. The same prompts had been run once before inside the project workspace for the
sensitivity panel of Section~\ref{sec:memory-panel} (306 calls including 8 aborted,
12{,}572{,}889 input and 1{,}015{,}945 output tokens). An earlier re-collection with batches of 90 requested 2{,}547 judgments in 31 batches. The version-pinned re-collection made 669 calls: 212 each for the opus and sonnet tiers (112 batched and 100 single-case, no retry) and 245 for haiku, of which 39 were retries, stops by a platform filter or empty transcripts. It used 114{,}268{,}325 input tokens, of which 95{,}492{,}861 were cache reads, and 585{,}358 output tokens. The third-party corpus run made 548 calls: 178 batched and 351 single-case, which include 53 retries, and 19 that a platform filter stopped without a verdict, all in the haiku tier. It used 84{,}702{,}072 input tokens, of which 71{,}058{,}077 were cache reads, and 447{,}205 output tokens. Another 13 probe and pilot calls are not counted; some of them were served by a version other than the one specified and were discarded. The internal corpus was not re-collected because it cannot distinguish
stack configurations (Section~\ref{sec:limitations-m4}, item 4). Analysis ran on
a single laptop.

%% file: sections/92_ethics.tex
\section*{Ethical Considerations}
The corpora consist of agent actions labelled dangerous or benign. The dangerous
cases are command and code strings of known categories (data exfiltration,
credential exposure, destructive operations and persistence) written as test
cases. None is a new exploit, and none targets a live system. The external corpus and the
prompts that embed the case texts are not part of the public artifact and are shared
on request; the internal and independent corpora were already public
\cite{agenttrust2026a}. No human subjects were involved
in this study. The paper names the judge models for reproducibility and draws no
conclusion about any vendor. The verdict-accounting finding of
Section~\ref{sec:verdict-alphabet} concerns an evaluation convention.

%% file: sections/99_appendix.tex
\section{Supplementary Tables}\label{app:tables}

\begin{table}[h]
\caption{Per-corpus miss correlation (PRIMARY): pooled odds ratio, the
Cochran--Mantel--Haenszel common odds ratio stratified by corpus with its $p$-value,
and $\varphi$ within each corpus. A Breslow--Day test of homogeneity of the
per-corpus odds ratios gives $p \ge 0.064$ for every pair except gpt55 with opus
($p = 0.026$).}
\label{tab:stratified}
\centering
\scriptsize
\setlength{\tabcolsep}{3pt}
\begin{tabular}{@{}lrrrrrr@{}}
\toprule
pair & pooled OR & MH OR & MH $p$ & int. & ind. & ext. \\
\midrule
rules $\times$ gpt55 & 2.19 & 3.96 & 0.50 & $-0.02$ & $+0.06$ & $+0.10$ \\
rules $\times$ haiku & 1.19 & 2.10 & 0.12 & $-0.08$ & $+0.16$ & $-0.01$ \\
rules $\times$ opus & 0.86 & 2.18 & 0.17 & $-0.08$ & $+0.11$ & $+0.10$ \\
rules $\times$ sonnet & 0.72 & 1.74 & 0.30 & $-0.09$ & $+0.07$ & $+0.18$ \\
gpt55 $\times$ haiku & 28.25 & 26.68 & $2\times10^{-7}$ & $-0.03$ & $+0.28$ & $+0.49$ \\
gpt55 $\times$ opus & 28.25 & 25.51 & $2\times10^{-7}$ & $-0.03$ & $+0.27$ & $+1.00$ \\
gpt55 $\times$ sonnet & 8.56 & 7.48 & $5\times10^{-3}$ & $-0.03$ & $+0.14$ & $+0.57$ \\
haiku $\times$ opus & 15.18 & 12.82 & $2\times10^{-20}$ & $+0.60$ & $+0.38$ & $+0.49$ \\
haiku $\times$ sonnet & 23.21 & 21.23 & $8\times10^{-29}$ & $+0.67$ & $+0.48$ & $+0.26$ \\
opus $\times$ sonnet & 48.47 & 43.43 & $9\times10^{-38}$ & $+0.50$ & $+0.62$ & $+0.57$ \\
\bottomrule
\end{tabular}
\end{table}

\begin{table}[h]
\caption{$\nmult$ of every two-layer stack by corpus (STRICT), with the exact 95\%
interval and the floor under perfect coupling. A censored stack (no joint miss) is
given as its lower bound with an open upper limit.}
\label{tab:nmult-corpus}
\centering
\scriptsize
\setlength{\tabcolsep}{2pt}
\begin{tabular}{@{}lllllll@{}}
\toprule
 & \multicolumn{2}{c}{internal (129)} & \multicolumn{2}{c}{independent (344)} & \multicolumn{2}{c}{external (87)} \\
stack & $\nmult$ [95\%] & floor & $\nmult$ [95\%] & floor & $\nmult$ [95\%] & floor \\
\midrule
rules + gpt55 & $\ge 1.42$ & 1.06 & 1.82 [1.41, 2.43] & 1.07 & 1.96 [1.43, 2.65] & 1.54 \\
rules + haiku & $\ge 1.50$ & 1.12 & 1.66 [1.35, 2.04] & 1.19 & 1.76 [1.29, 2.38] & 1.58 \\
rules + opus & $\ge 1.39$ & 1.04 & 1.70 [1.38, 2.12] & 1.17 & 2.05 [1.27, 3.73] & 1.73 \\
rules + sonnet & $\ge 1.56$ & 1.16 & 1.72 [1.43, 2.09] & 1.32 & 1.75 [1.25, 2.43] & 1.64 \\
gpt55 + haiku & 1.31 [1.00, 1.71] & 1.06 & 1.37 [1.16, 1.61] & 1.12 & 1.26 [0.91, 1.72] & 1.07 \\
gpt55 + opus & 1.20 [0.92, 1.58] & 1.02 & 1.35 [1.15, 1.58] & 1.10 & 1.31 [0.88, 2.04] & 1.31 \\
gpt55 + sonnet & 1.24 [0.96, 1.60] & 1.11 & 1.46 [1.24, 1.70] & 1.25 & 1.62 [1.08, 2.53] & 1.15 \\
haiku + opus & 1.22 [0.94, 1.59] & 1.08 & 1.33 [1.14, 1.55] & 1.02 & 1.48 [0.92, 2.70] & 1.25 \\
haiku + sonnet & 1.23 [0.96, 1.57] & 1.04 & 1.32 [1.14, 1.52] & 1.14 & 1.36 [0.94, 2.00] & 1.08 \\
opus + sonnet & 1.17 [0.91, 1.50] & 1.12 & 1.24 [1.07, 1.42] & 1.16 & 1.39 [0.86, 2.53] & 1.17 \\
\bottomrule
\end{tabular}
\end{table}

\begin{table}[h]
\caption{Per-corpus medians of $\varphi$ (PRIMARY) for the two groups of pairs.}
\label{tab:percorpus-phi}
\centering
\footnotesize
\begin{tabular}{@{}lll@{}}
\toprule
corpus & rules $\times$ judge & judge $\times$ judge \\
\midrule
internal & $-0.079$ & $+0.234$ \\
independent & $+0.089$ & $+0.331$ \\
external (held out) & $+0.102$ & $+0.531$ \\
\bottomrule
\end{tabular}
\end{table}

\begin{table}[h]
\caption{Equivalence testing on the rule pairs: exact 90\% odds-ratio intervals.
The pre-registered equivalence region was $[0.50, 2.00]$.}
\label{tab:tost}
\centering
\footnotesize
\begin{tabular}{@{}lll@{}}
\toprule
pair & PRIMARY & STRICT \\
\midrule
rules $\times$ gpt55 & [0.31, 10.02] & [0.70, 2.74] \\
rules $\times$ haiku & [0.56, 2.32] & [0.81, 2.61] \\
rules $\times$ opus & [0.37, 1.79] & [0.34, 1.62] \\
rules $\times$ sonnet & [0.36, 1.35] & [0.51, 1.59] \\
\bottomrule
\end{tabular}
\end{table}

\subsection{Conclusions withdrawn during analysis}
Seven candidate conclusions were stated during analysis and withdrawn. Each is
listed with the check that removed it.
\begin{enumerate}
\item ``The layer that adds the most to the stack is the one with the worst solo
  accuracy.'' Falsified by the second rule engine, b1, which has the worst solo
  accuracy and adds 9 cases (Section~\ref{sec:instrument-fails}).
\item ``Rule-plus-judge stacks beat judge-plus-judge stacks'' as a headline. In the
  stratified replication the held-out corpus favoured the rule stack in 0 of 12
  comparisons under PRIMARY (6 tied) and 1 of 12 under STRICT, and 4 of 12 and 9 of
  12 comparisons reversed direction between
  corpora. The
  comparison measures independence times competence
  (Section~\ref{sec:decoupling-not-usefulness}).
\item ``The rule layer and the LLM layer are independent.'' Equivalence testing
  failed at both pre-registered bounds (Section~\ref{sec:not-established}).
\item ``Model version changes concentrate on the most sensitive batches.'' Two
  events, one of them a counterexample. The densest segment of the external run was unaffected. On the primary panel the
  change was associated with the corpus (Section~\ref{sec:version-changes}); no
  mechanism claim is made.
\item ``Cross-vendor judge pairs recover some independence.'' On the primary panel the
  archived cross-vendor pairs' excess over the same-vendor pairs halves under STRICT
  and is confounded with that column's collection, and the cross-vendor pairs with
  clean provenance read inside the same-vendor band
  (Sections~\ref{sec:readings} and~\ref{sec:cross-vendor-arm}).
\item ``The difficulty share has a lower bound near 47\%.'' The computation that
  produced it scored difficulty with a probe that admitted the archived faithful
  DeepSeek column, a column excluded from every other headline number, and the
  share moves by a factor of two with the probe. Withdrawn after review in favour of the range in
  Section~\ref{sec:where-coupling-comes-from}.
\item ``The re-collected judges saw no mention of this study.'' A probe of the
  subagent system context showed a project-memory index naming the runtime and the
  venue; the claim is withdrawn, the affected collection is retained as a comparison
  panel, and the primary panel was collected in a session without that index
  (Section~\ref{sec:memory-panel}).
\end{enumerate}

\subsection{A false-alarm observation on the cross-vendor column}
On the 246 benign cases the rule layer and the fresh DeepSeek column false-alarm
together ($\varphi = +0.305$, Holm $p = 0.004$ over the fifteen pairs of the
six-layer panel, with 6 joint false alarms out of 17 and
17). The archived DeepSeek column
shares 8 of the rule layer's 17. The other four judges share 0 to 1. All six joint
cases are external-corpus cases in the network-egress, observability and CI
categories: benign commands that look like exfiltration or probing. On the
false-alarm axis this one judge behaves partly like a surface-pattern detector. We
record it as an observation about one model family at $n = 6$ and do not build on
it. The miss-side split of Section~\ref{sec:readings} is unaffected; the observation sits here because it rests on six cases.

\subsection{Verification coverage}
Six headline statistics were recomputed against an independent implementation. The
Cochran--Mantel--Haenszel common odds ratio was compared with a statistics library
on all ten pairs and agreed exactly; $\varphi$ with Pearson's $r$ on the 0/1 error
vectors, agreeing to within $2.6 \times 10^{-15}$; Cohen's $\kappa$ with a
separately written confusion-matrix formula, to within $1.8 \times 10^{-15}$; the
exact McNemar tests with a library implementation on all twelve comparisons,
exactly; $\nmult$ with a hand calculation from its definition, exactly; and the
difficulty share with a second implementation written after review, which agrees with the released script under the same probe construction (31.8\%
against 32\%).
Holm correction and the new-coverage counts were also recomputed by a second method
with exact agreement. The rule column was regenerated under the archived,
description-visible condition and matched the archived column on all 1{,}119
cases.

%% file: references.bib
@article{agenttrust2026a,
  title   = {{AgentTrust}: Runtime Safety Evaluation and Interception for
             {AI} Agent Tool Use},
  author  = {Yang, Chenglin},
  journal = {arXiv preprint arXiv:2605.04785},
  year    = {2026}
}

@article{agenttrust2026b,
  title   = {{AgentTrust}: A Self-Improving Trust Layer for {AI}-Agent Actions},
  author  = {Yang, Chenglin},
  journal = {arXiv preprint arXiv:2606.08539},
  year    = {2026}
}

@article{ray2026enforced,
  title   = {What Can Be Enforced? {A} Theory of Certified Runtime Safety for
             Tool-Using Agents},
  author  = {Ray, Shawn},
  journal = {arXiv preprint arXiv:2607.22868},
  year    = {2026}
}

@article{zhuang2026agenttrap,
  title   = {{AgentTrap}: Measuring Runtime Trust Failures in Third-Party
             Agent Skills},
  author  = {Zhuang, Haomin and Xing, Hanwen and Zhou, Yujun and Ma, Yuchen
             and Huang, Yue and Shen, Yili and Han, Yufei and
             Zhang, Xiangliang},
  journal = {arXiv preprint arXiv:2605.13940},
  year    = {2026}
}

@article{chennabasappa2025llamafirewall,
  title   = {{LlamaFirewall}: An Open Source Guardrail System for Building
             Secure {AI} Agents},
  author  = {Chennabasappa, Sahana and Nikolaidis, Cyrus and Song, Daniel and
             Molnar, David and Ding, Stephanie and Wan, Shengye and
             Whitman, Spencer and Deason, Lauren and Doucette, Nicholas and
             Montilla, Abraham and Gampa, Alekhya and de Paola, Beto and
             Gabi, Dominik and Crnkovich, James and Testud, Jean-Christophe
             and He, Kat and Chaturvedi, Rashnil and Zhou, Wu and
             Saxe, Joshua},
  journal = {arXiv preprint arXiv:2505.03574},
  year    = {2025}
}

@article{inan2023llamaguard,
  title   = {{Llama Guard}: {LLM}-Based Input-Output Safeguard for Human-{AI}
             Conversations},
  author  = {Inan, Hakan and Upasani, Kartikeya and Chi, Jianfeng and
             Rungta, Rashi and Iyer, Krithika and Mao, Yuning and
             Tontchev, Michael and Hu, Qing and Fuller, Brian and
             Testuggine, Davide and Khabsa, Madian},
  journal = {arXiv preprint arXiv:2312.06674},
  year    = {2023}
}

@inproceedings{rebedea2023nemo,
  title     = {{NeMo Guardrails}: A Toolkit for Controllable and Safe {LLM}
               Applications with Programmable Rails},
  author    = {Rebedea, Traian and Dinu, Razvan and Sreedhar, Makesh Narsimhan and
               Parisien, Christopher and Cohen, Jonathan},
  booktitle = {EMNLP System Demonstrations},
  pages     = {431--445},
  year      = {2023},
  doi      = {10.18653/v1/2023.emnlp-demo.40}
}

@inproceedings{greshake2023injection,
  title     = {Not What You've Signed Up For: Compromising Real-World
               {LLM}-Integrated Applications with Indirect Prompt Injection},
  author    = {Greshake, Kai and Abdelnabi, Sahar and Mishra, Shailesh and
               Endres, Christoph and Holz, Thorsten and Fritz, Mario},
  booktitle = {ACM Workshop on Artificial Intelligence and Security (AISec)},
  year      = {2023},
  doi      = {10.1145/3605764.3623985},
  pages    = {79--90}
}

@inproceedings{debenedetti2024agentdojo,
  title     = {{AgentDojo}: A Dynamic Environment to Evaluate Prompt Injection
               Attacks and Defenses for {LLM} Agents},
  author    = {Debenedetti, Edoardo and Zhang, Jie and Balunovi{\'c}, Mislav
               and Beurer-Kellner, Luca and Fischer, Marc and
               Tram{\`e}r, Florian},
  booktitle = {NeurIPS Datasets and Benchmarks Track},
  year      = {2024},
  doi      = {10.52202/079017-2636}
}

@article{andriushchenko2024agentharm,
  title   = {{AgentHarm}: A Benchmark for Measuring Harmfulness of {LLM} Agents},
  author  = {Andriushchenko, Maksym and Souly, Alexandra and Dziemian, Mateusz
             and Duenas, Derek and Lin, Maxwell and Wang, Justin and
             Hendrycks, Dan and Zou, Andy and Kolter, Zico and
             Fredrikson, Matt and Winsor, Eric and Wynne, Jerome and
             Gal, Yarin and Davies, Xander},
  journal = {arXiv preprint arXiv:2410.09024},
  year    = {2024}
}

@inproceedings{ruan2024toolemu,
  title     = {Identifying the Risks of {LM} Agents with an {LM}-Emulated Sandbox},
  author    = {Ruan, Yangjun and Dong, Honghua and Wang, Andrew and Pitis, Silviu
               and Zhou, Yongchao and Ba, Jimmy and Dubois, Yann and
               Maddison, Chris J. and Hashimoto, Tatsunori},
  booktitle = {International Conference on Learning Representations (ICLR)},
  year      = {2024}
}

@inproceedings{yuan2024rjudge,
  title     = {{R-Judge}: Benchmarking Safety Risk Awareness for {LLM} Agents},
  author    = {Yuan, Tongxin and He, Zhiwei and Dong, Lingzhong and Wang, Yiming
               and Zhao, Ruijie and Xia, Tian and Xu, Lizhen and Zhou, Binglin
               and Li, Fangqi and Zhang, Zhuosheng and Wang, Rui and
               Liu, Gongshen},
  booktitle = {Findings of EMNLP},
  year      = {2024},
  doi      = {10.18653/v1/2024.findings-emnlp.79},
  pages    = {1467--1490}
}

@article{li2025robustjudge,
  title   = {{LLMs} Cannot Reliably Judge (Yet?): A Comprehensive Assessment on the
             Robustness of {LLM}-as-a-Judge},
  author  = {Li, Songze and Xu, Chuokun and Wang, Jiaying and Gong, Xueluan and
             Chen, Chen and Zhang, Jirui and Wang, Jun and Lam, Kwok-Yan and
             Ji, Shouling},
  journal = {arXiv preprint arXiv:2506.09443},
  year    = {2025},
}

@article{cohen1960kappa,
  title   = {A Coefficient of Agreement for Nominal Scales},
  author  = {Cohen, Jacob},
  journal = {Educational and Psychological Measurement},
  volume  = {20},
  number  = {1},
  pages   = {37--46},
  year    = {1960},
  doi      = {10.1177/001316446002000104}
}

@misc{register2025replit,
  title        = {Vibe Coding Service {Replit} Deleted User's Production
                  Database, Faked Data, Told Fibs Galore},
  author       = {Sharwood, Simon},
  year         = {2025},
  note         = {\emph{The Register}, 21 July 2025},
  howpublished = {\url{https://www.theregister.com/2025/07/21/replit_saastr_vibe_coding_incident/}}
}

@article{alotaibi2026layered,
  title   = {Layered {LLM} Defenses as an Ensemble: Access Tiers, Inference Cost,
             and the Measured Failure Correlation Between Defense Layers},
  author  = {Alotaibi, Abrar and Jabbar, Muhammad Shahid and Al-Azani, Sadam and
             Ahmed, Moataz},
  journal = {arXiv preprint arXiv:2608.28327},
  year    = {2026},
}

@article{fokou2026parallax,
  title   = {Parallax: Why {AI} Agents That Think Must Never Act},
  author  = {Fokou, Joel},
  journal = {arXiv preprint arXiv:2604.12986},
  year    = {2026},
}

@article{debenedetti2025camel,
  title   = {Defeating Prompt Injections by Design},
  author  = {Debenedetti, Edoardo and Shumailov, Ilia and Fan, Tianqi and
             Hayes, Jamie and Carlini, Nicholas and Fabian, Daniel and
             Kern, Christoph and Shi, Chongyang and Terzis, Andreas and
             Tram{\`e}r, Florian},
  journal = {arXiv preprint arXiv:2503.18813},
  year    = {2025},
}

@article{costa2025fides,
  title   = {Securing {AI} Agents with Information-Flow Control},
  author  = {Costa, Manuel and K{\"o}pf, Boris and Kolluri, Aashish and
             Paverd, Andrew and Russinovich, Mark and Salem, Ahmed and
             Tople, Shruti and Wutschitz, Lukas and Zanella-B{\'e}guelin, Santiago},
  journal = {arXiv preprint arXiv:2505.23643},
  year    = {2025},
}

@article{shi2025progent,
  title   = {Progent: Securing {AI} Agents with Privilege Control},
  author  = {Shi, Tianneng and He, Jingxuan and Wang, Zhun and Li, Hongwei and
             Wu, Linyu and Guo, Wenbo and Song, Dawn},
  journal = {arXiv preprint arXiv:2504.11703},
  year    = {2025},
}

@article{kohli2026ninejudges,
  title   = {Nine Judges, Two Effective Votes: Correlated Errors Undermine {LLM}
             Evaluation Panels},
  author  = {Kohli, Guneet},
  journal = {arXiv preprint arXiv:2605.29800},
  year    = {2026},
}

@book{kish1965survey,
  title     = {Survey Sampling},
  author    = {Kish, Leslie},
  publisher = {John Wiley \& Sons},
  address   = {New York},
  year      = {1965},
}

@article{hill1973diversity,
  title   = {Diversity and Evenness: A Unifying Notation and Its Consequences},
  author  = {Hill, Mark O.},
  journal = {Ecology},
  volume  = {54},
  number  = {2},
  pages   = {427--432},
  year    = {1973},
  doi      = {10.2307/1934352}
}

@article{macarthur1965patterns,
  title   = {Patterns of Species Diversity},
  author  = {MacArthur, Robert H.},
  journal = {Biological Reviews},
  volume  = {40},
  number  = {4},
  pages   = {510--533},
  year    = {1965},
  doi      = {10.1111/j.1469-185X.1965.tb00815.x}
}

@book{macarthur1972geographical,
  title     = {Geographical Ecology: Patterns in the Distribution of Species},
  author    = {MacArthur, Robert H.},
  publisher = {Harper \& Row},
  year      = {1972},
}

@article{jost2006entropy,
  title   = {Entropy and Diversity},
  author  = {Jost, Lou},
  journal = {Oikos},
  volume  = {113},
  number  = {2},
  pages   = {363--375},
  year    = {2006},
  doi      = {10.1111/j.2006.0030-1299.14714.x}
}

@article{jost2007partitioning,
  title   = {Partitioning Diversity into Independent Alpha and Beta Components},
  author  = {Jost, Lou},
  journal = {Ecology},
  volume  = {88},
  number  = {10},
  pages   = {2427--2439},
  year    = {2007},
  doi      = {10.1890/06-1736.1}
}

@techreport{iec61508,
  title       = {{IEC} 61508: Functional Safety of Electrical/Electronic/Programmable
                 Electronic Safety-Related Systems},
  author      = {{International Electrotechnical Commission}},
  institution = {IEC},
  address     = {Geneva},
  year        = {2010},
}

@article{eckhardt1985theoretical,
  title   = {A Theoretical Basis for the Analysis of Multiversion Software Subject
             to Coincident Errors},
  author  = {Eckhardt, Dave E. and Lee, Larry D.},
  journal = {IEEE Transactions on Software Engineering},
  volume  = {SE-11},
  number  = {12},
  pages   = {1511--1517},
  year    = {1985},
  doi     = {10.1109/TSE.1985.231895}
}

@article{knight1986experimental,
  title   = {An Experimental Evaluation of the Assumption of Independence in
             Multiversion Programming},
  author  = {Knight, John C. and Leveson, Nancy G.},
  journal = {IEEE Transactions on Software Engineering},
  volume  = {SE-12},
  number  = {1},
  pages   = {96--109},
  year    = {1986},
  doi     = {10.1109/TSE.1986.6312924}
}

@article{littlewood1989conceptual,
  title   = {Conceptual Modeling of Coincident Failures in Multiversion Software},
  author  = {Littlewood, Bev and Miller, Douglas R.},
  journal = {IEEE Transactions on Software Engineering},
  volume  = {15},
  number  = {12},
  pages   = {1596--1614},
  year    = {1989},
  doi     = {10.1109/32.58771}
}

@article{kuncheva2003diversity,
  title   = {Measures of Diversity in Classifier Ensembles and Their Relationship
             with the Ensemble Accuracy},
  author  = {Kuncheva, Ludmila I. and Whitaker, Christopher J.},
  journal = {Machine Learning},
  volume  = {51},
  number  = {2},
  pages   = {181--207},
  year    = {2003},
  doi      = {10.1023/A:1022859003006}
}

@article{holm1979simple,
  title   = {A Simple Sequentially Rejective Multiple Test Procedure},
  author  = {Holm, Sture},
  journal = {Scandinavian Journal of Statistics},
  volume  = {6},
  number  = {2},
  pages   = {65--70},
  year    = {1979},
}

@article{mantel1959statistical,
  title   = {Statistical Aspects of the Analysis of Data from Retrospective Studies
             of Disease},
  author  = {Mantel, Nathan and Haenszel, William},
  journal = {Journal of the National Cancer Institute},
  volume  = {22},
  number  = {4},
  pages   = {719--748},
  year    = {1959},
  doi      = {10.1093/jnci/22.4.719}
}

@article{clopper1934confidence,
  title   = {The Use of Confidence or Fiducial Limits Illustrated in the Case of
             the Binomial},
  author  = {Clopper, Charles J. and Pearson, Egon S.},
  journal = {Biometrika},
  volume  = {26},
  number  = {4},
  pages   = {404--413},
  year    = {1934},
  doi      = {10.1093/biomet/26.4.404}
}

@article{mcnemar1947note,
  title   = {Note on the Sampling Error of the Difference Between Correlated
             Proportions or Percentages},
  author  = {McNemar, Quinn},
  journal = {Psychometrika},
  volume  = {12},
  number  = {2},
  pages   = {153--157},
  year    = {1947},
  doi      = {10.1007/BF02295996}
}

@article{turan2026oversight,
  title   = {Oversight Has a Capacity: Calibrating Agent Guards to a Subjective,
             Fatiguing Human},
  author  = {Turan, Emre},
  journal = {arXiv preprint arXiv:2606.08919},
  year    = {2026},
}

@article{goel2025alike,
  title   = {Great Models Think Alike and this Undermines {AI} Oversight},
  author  = {Goel, Shashwat and Struber, Joschka and Auzina, Ilze Amanda and
             Chandra, Karuna K and Kumaraguru, Ponnurangam and Kiela, Douwe and
             Prabhu, Ameya and Bethge, Matthias and Geiping, Jonas},
  journal = {arXiv preprint arXiv:2502.04313},
  year    = {2025},
}

@article{kim2025correlated,
  title   = {Correlated Errors in Large Language Models},
  author  = {Kim, Elliot and Garg, Avi and Peng, Kenny and Garg, Nikhil},
  journal = {arXiv preprint arXiv:2506.07962},
  year    = {2025},
}

@article{nasr2025attacker,
  title   = {The Attacker Moves Second: Stronger Adaptive Attacks Bypass Defenses
             Against {LLM} Jailbreaks and Prompt Injections},
  author  = {Nasr, Milad and Carlini, Nicholas and Sitawarin, Chawin and
             Schulhoff, Sander V. and Hayes, Jamie and Ilie, Michael and
             Pluto, Juliette and Song, Shuang and Chaudhari, Harsh and
             Shumailov, Ilia and Thakurta, Abhradeep and Xiao, Kai Yuanqing and
             Terzis, Andreas and Tram{\`e}r, Florian},
  journal = {arXiv preprint arXiv:2510.09023},
  year    = {2025},
}

@article{zhan2025adaptive,
  title   = {Adaptive Attacks Break Defenses Against Indirect Prompt Injection
             Attacks on {LLM} Agents},
  author  = {Zhan, Qiusi and Fang, Richard and Panchal, Henil Shalin and Kang, Daniel},
  journal = {arXiv preprint arXiv:2503.00061},
  year    = {2025},
}

@inproceedings{bishop2011diversity,
  title     = {Diversity for Security: A Study with Off-the-Shelf {AntiVirus} Engines},
  author    = {Bishop, Peter and Bloomfield, Robin and Gashi, Ilir and Stankovic, Vladimir},
  booktitle = {IEEE 22nd International Symposium on Software Reliability Engineering (ISSRE)},
  pages     = {11--19},
  year      = {2011},
  doi       = {10.1109/ISSRE.2011.15},
}

@inproceedings{gashi2013malware,
  title     = {Does Malware Detection Improve with Diverse {AntiVirus} Products?
               An Empirical Study},
  author    = {Gashi, Ilir and Sobesto, Bertrand and Stankovic, Vladimir and Cukier, Michel},
  booktitle = {Computer Safety, Reliability, and Security (SAFECOMP), LNCS},
  pages     = {94--105},
  year      = {2013},
  doi       = {10.1007/978-3-642-40793-2_9},
}

@article{littlewood2001diversity,
  title   = {Modeling Software Design Diversity: A Review},
  author  = {Littlewood, Bev and Popov, Peter and Strigini, Lorenzo},
  journal = {ACM Computing Surveys},
  volume  = {33},
  number  = {2},
  pages   = {177--208},
  year    = {2001},
  doi     = {10.1145/384192.384195},
}

@inproceedings{guo2024redcode,
  title     = {{RedCode}: Risky Code Execution and Generation Benchmark for Code
               Agents},
  author    = {Guo, Chengquan and Liu, Xun and Xie, Chulin and Zhou, Andy and Zeng,
               Yi and Lin, Zinan and Song, Dawn and Li, Bo},
  booktitle = {Advances in Neural Information Processing Systems (NeurIPS),
               Datasets and Benchmarks Track},
  year      = {2024}
}
